%% file: main.tex
\documentclass[10pt,twocolumn,letterpaper]{article}

\usepackage[accsupp]{axessibility}
\usepackage{cvpr}              % To produce the CAMERA-READY version
\usepackage{booktabs} 
\usepackage{multirow}
\usepackage{gensymb}
\usepackage{tcolorbox} % 导入 tcolorbox 包
\usepackage[ruled,vlined]{algorithm2e}
\usepackage{dblfloatfix}
\usepackage{caption}
\usepackage{etoolbox}
\usepackage[dvipsnames]{xcolor}
\usepackage{color}
\usepackage{CJKutf8}
\usepackage{lineno}
\usepackage[accsupp]{axessibility}
\definecolor{cvprblue}{rgb}{0.21,0.49,0.74}
\usepackage[pagebackref,breaklinks,colorlinks,allcolors=cvprblue]{hyperref}

\def\paperID{1878} % *** Enter the Paper ID here
\def\confName{CVPR}
\def\confYear{2025}

\title{Towards Long-Horizon Vision-Language Navigation: \\ Platform, Benchmark and Method}

\author{Xinshuai Song\\
Sun Yat-sen University\\
{\tt\small songxsh@mail2.sysu.edu.cn}
\and
Weixing Chen\\
Sun Yat-sen University\\
{\tt\small chenwx228@mail2.sysu.edu.cn}
\and
Yang Liu\\
Sun Yat-sen University\\
{\tt\small liuy856@mail.sysu.edu.cn}
\and
Weikai Chen\\
Tencent America\\
{\tt\small chenwk891@gmail.com}
\and
Guanbin Li\\
Sun Yat-sen University\\
{\tt\small liguanbin@mail.sysu.edu.cn}
\and
Liang Lin\\
Sun Yat-sen University\\
{\tt\small linliang@ieee.org}
}
\author{
Xinshuai Song$^1$\thanks{Equal contribution} \hspace{0.7em} Weixing Chen$^{1*}$ \hspace{0.7em} Yang Liu$^{1,3}$\thanks{Corresponding Author} \hspace{0.7em}  Weikai Chen\thanks{This paper solely reflects the author's personal research and is not associated with the author's affiliated institution.} 
\hspace{0.7em}  Guanbin Li$^{1,2,3}$ \hspace{0.7em}  Liang Lin$^{1,2,3}$\\
$^1$Sun Yat-sen University, China $^2$Peng Cheng Laboratory \\$^3$Guangdong Key Laboratory of Big Data Analysis and Processing\\
{\tt\small \{songxsh,chenwx228\}@mail2.sysu.edu.cn,liuy856@mail.sysu.edu.cn,chenwk891@gmail.com}\\
{\tt\small liguanbin@mail.sysu.edu.cn,linliang@ieee.org}\\
{\tt\small \href{https://hcplab-sysu.github.io/LH-VLN}{hcplab-sysu.github.io/LH-VLN}}
}

\makeatletter
\patchcmd{\@maketitle}
  {\end{center}}
  {%
    \end{center}%
    \vspace{-3em}%
    \begin{center}%
      \includegraphics[width=0.86\linewidth]{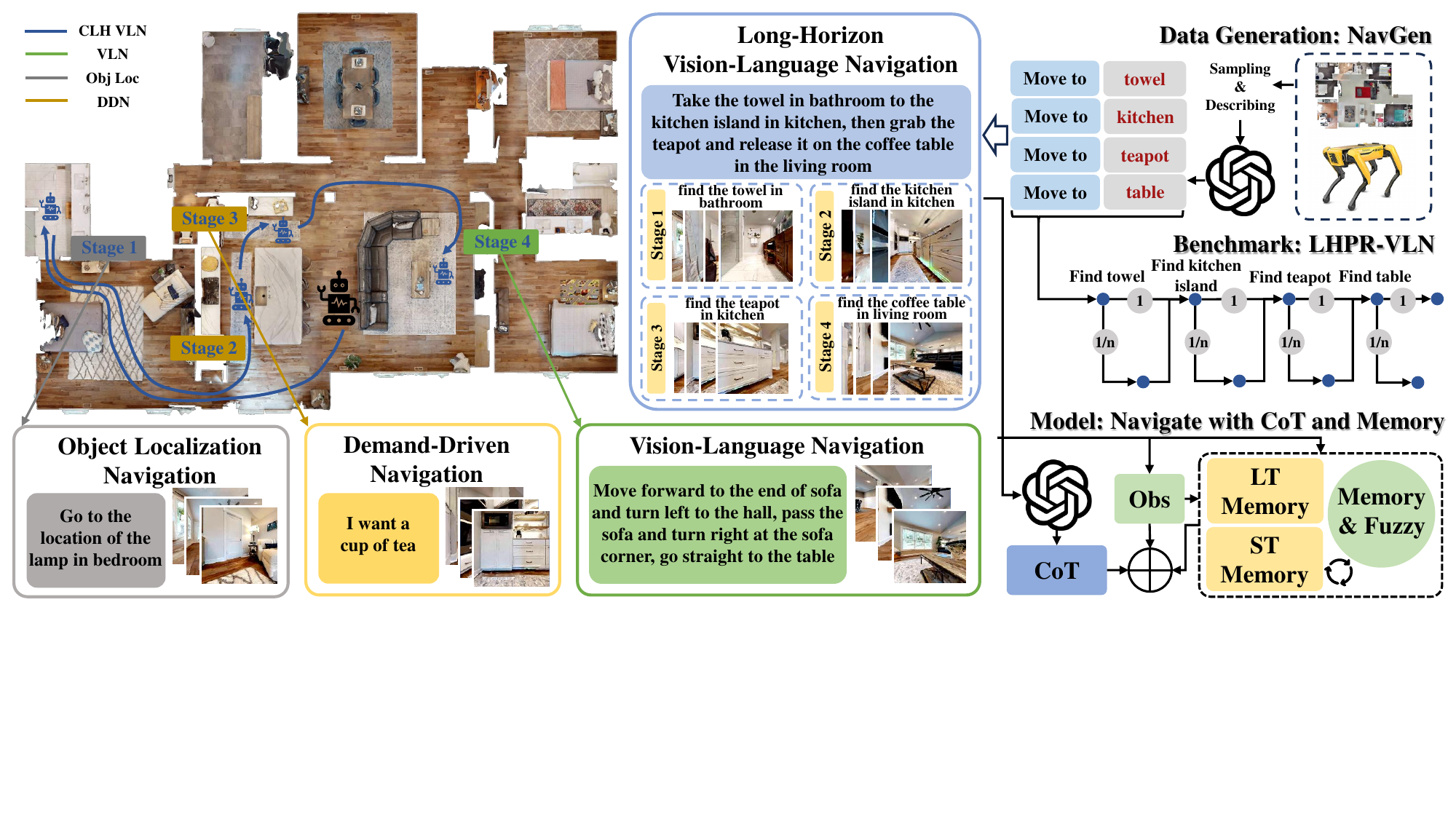}\par%
      \vspace{-1em}%
      \captionof{figure}{Framework Overview. Different from existing vision language navigation, object loco-navigation, and demand-driven navigation benchmarks, LH-VLN divides navigation into multiple subtasks, requiring the agent to complete these tasks sequentially within the scene. Our data generation framework provides a general LH-VLN task generation pipeline, and the newly built LHPR-VLN benchmark for multi-stage navigation tasks. Our navigation model, based on the chain-of-thought (CoT) feedback and adaptive memory design, achieves efficient navigation by utilizing CoT prompts and dynamic long-term and short-term memories. }%
      \label{fig:intro}
      \vspace{0em}%
    \end{center}%
  }
  {}{}
\makeatother

\begin{document}
\maketitle

% % \twocolumn[{%
% % % \renewcommand\twocolumn[1][]{#1}%
% % \maketitle
% % \centering
% % \includegraphics[width=0.88\linewidth]{sec/img/1-intro.pdf}
% %  \vspace{-1em}
% % \captionof{figure}{Framework Overview. Different from existing vision language navigation, object loco-navigation, and demand-driven navigation benchmarks, LH-VLN divides navigation into multiple subtasks, requiring the agent to complete these tasks sequentially within the scene. Our data generation framework provides a general LH-VLN task generation pipeline, and the newly built LHPR-VLN benchmark for multi-stage navigation tasks. Our navigation model, based on the chain-of-thought (CoT) feedback and adaptive memory design, achieves efficient navigation by utilizing CoT prompts and dynamic long-term and short-term memories. }
% %  \vspace{0.5em}
% %     \label{fig:intro}
% % }]

\input{sec/0_abstract}    
\input{sec/1_introduction}
\input{sec/2_related_work}

\input{sec/3_muti_stage_causal_navigation_task}
\input{sec/4_method}
\input{sec/5_experiment}
\input{sec/6_conclusion}

\input{sec/X_suppl}

\clearpage
{
    \small
    \bibliographystyle{ieeenat_fullname}
    \bibliography{main}
}


\end{document}

%% file: sec/0_abstract.tex
\begin{abstract}
% 动机
%Advancements in Multi-modal Large Models (MLMs) have propelled Vision-Language Navigation (VLN) closer to real-world applications. 

Existing Vision-Language Navigation (VLN) methods primarily focus on single-stage navigation, limiting their effectiveness in multi-stage and long-horizon tasks within complex and dynamic environments. 
% 提出任务
To address these limitations, we propose a novel VLN task, named Long-Horizon Vision-Language Navigation (LH-VLN), which emphasizes long-term planning and decision consistency across consecutive subtasks. 
% 数据构建平台
Furthermore, to support LH-VLN, we develop an automated data generation platform NavGen, which constructs datasets with complex task structures and improves data utility through a bidirectional, multi-granularity generation approach. 
% 基准
To accurately evaluate complex tasks, we construct the Long-Horizon Planning and Reasoning in VLN (LHPR-VLN) benchmark consisting of 3,260 tasks with an average of 150 task steps, serving
as the first dataset specifically designed for the long-horizon vision-language navigation task. Furthermore, we propose Independent Success Rate (ISR), Conditional Success Rate (CSR), and CSR weight by Ground Truth (CGT) metrics, to provide fine-grained assessments of task completion. 
% 方法
To improve model adaptability in complex tasks, we propose a novel Multi-Granularity Dynamic Memory (MGDM) module that integrates short-term memory blurring with long-term memory retrieval to enable flexible navigation in dynamic environments. Our platform, benchmark and method supply LH-VLN with a robust data generation pipeline, comprehensive model evaluation dataset, reasonable metrics, and a novel VLN model, establishing a foundational framework for advancing LH-VLN. 
\end{abstract}

%% file: sec/1_introduction.tex
\section{Introduction}
\label{sec:introduction}

% 缺少数据集 -> 建立数据生成平台：数据利用率低 -> 双向生成： 难以迁移 -> 通用平台
% 当前基准不合适LH-VLN -> 建立多粒度的评估基准
% 当前方法对记忆的处理过于粗暴 -> 自适应的动态记忆模块

% 面对真实世界场景，当前的方法在动态环境与多阶段任务存在局限。这种局限来自于缺少长程规划的能力，这是很难从单阶段任务中训练获得的。长程规划能力包括了，在一系列子任务的决策中保持连贯性以及在动态的反馈中纠错。

% \note{第一段应该是引出长程规划对于导航任务的重要性，写得需要更明确一些。}

Current Vision-Language Navigation (VLN) benchmarks and methods primarily focus on single-stage or short-term tasks, which involve simple objectives and limited action sequences, making them suitable for controlled settings but insufficient for real-world applications~\cite{liu2024aligning} (see Figure~\ref{fig:intro}).
In practical scenarios, agents must handle complex, long-horizon instructions that span multiple sub-tasks, requiring ongoing decision-making, dynamic re-planning, and sustained reasoning across extended periods~\cite{gu2022vision,mishra2023generative,sermanet2024robovqa}.
These capabilities are crucial for applications like autonomous assistants or service robots where coherent navigation over a long temporal horizon is essential. 
To address this gap, we propose, for the first time, a new task-coded Long-Horizon VLN (LH-VLN), to evaluate and enhance agents' abilities to manage multi-stage, context-rich navigation tasks that more accurately reflect real-world complexity.

The LH-VLN task is dedicated to push agents beyond simple, short-term navigation by requiring them to deeply comprehend complex task instructions, maintain continuous navigation, and handle sequential sub-tasks seamlessly across a dynamic environment.
Achieving this goal involves three critical components: 1)  an automated data generation platform to construct benchmarks with complex task structures and improves data utility, 2) a high-quality benchmark capturing the complexity of long-horizon, multi-stage tasks and accurately assess the agent’s task execution and detailed sub-task performance with reasonable metrics, and 3) a specialized method to equip agents with adaptive memory for complex navigation. 
In this work, we provide a comprehensive solution that addresses these three aspects, laying the foundation for robust LH-VLN in real-world scenarios.

%\note{现有的导航都是单阶段的吗？需要一句话过渡一下。}
% Existing VLN methods are typically modeled as single-stage tasks, where the agent receives a description of a single target including cues about objects, demands, and step-by-step instructions, then navigates accordingly~\cite{R2R_2018}. However, these tasks often involve short action sequences and relatively simple objectives, limiting their capacity to support complex multi-stage operations in real-world applications, as shown in Figure~\ref{fig:intro}. To this end, we propose the Long-Horizon VLN task (LH-VLN), which consists of a sequence of consecutive multi-stage sub-tasks. This task requires the agent to deeply comprehend task instructions and maintain continuous navigation within the environment. Furthermore, we provide a comprehensive solution for the proposed task due to the substantial gap between LH-VLN and current single-stage VLN tasks, including the lack of high-quality LH-VLN data, the inadequacy of benchmark evaluation methods for multi-stage sub-tasks, and the inability of naive memory-based models to meet the demands of complex scenarios. We propose an automated data generation platform, a dedicated multi-stage LH-VLN benchmark, and a reference model that moves beyond graph-memory~\cite{etpnav_2024,zhou2025navgpt} and semantic scene~\cite{wu2024embodied,long2024instructnav} strategies.

% \note{现有数据生成方法的缺点是什么？我们提出的数据生成方法命个名。}
Platform-wise, previous platforms \cite{kolve2017ai2,savva2019habitat,wang2023robogen, yang2024holodeck,wang2024towards} for VLN data generation lack sufficient versatility and depend on a specific simulation platform and assets, resulting in relatively limited generated data~\cite{deitke2023objaverse}. To overcome these limitations, we introduce \textit{NavGen}, a novel data generation platform that automates the construction of complex, multi-stage datasets. 
NavGen generates data through a bidirectional, multi-granularity approach, producing forward and backward sub-tasks to enrich task diversity and improve data utility. This automated platform allows for the scalable creation of richly varied navigation tasks that support advanced model training and long-horizon VLN evaluation.

% Given that current VLN datasets~\cite{R2R_2018, vln_ce_2020, REVERIE_2020, SOON_2021} suffer from limitations such as simple task structures, low data utilization, and limited flexibility in instructions, all of which constrain model generalization and hinder support for long-horizon tasks. Researchers often resort to labor-intensive manual annotation to compensate, significantly impeding progress in this field~\cite{wang2023robogen, yang2024holodeck}. 
% To address these issues, we propose a data generation platform NavGen, which automates dataset construction through three main steps: asset sampling, task generation, and trajectory generation. 
% % (1) sampling from a diverse set of 3D assets, including various scenes and robot models, to generate relevant descriptions; (2) using selected assets to further generate actions and task instructions; and (3) generating trajectory datasets based on these instructions. 
% To effectively boost data usability and improve model generalization, we introduce a bi-directional multi-granularity data generation method, which generates backward sub-tasks based on forward-generated trajectory data. These backward-generated tasks are related to the forward-generated ones but introduce fine-grained variations. 

Benchmark-wise, existing VLN benchmarks \cite{vln_ce_2020,SOON_2021,BEHAVIOR-1K_2023} are limited by their simple task structures, low data diversity, and constrained instructional flexibility, which restrict model generalization and hinder support for complex, long-horizon tasks. 
These benchmarks often rely on manual annotation, making them labor-intensive to create and less scalable for handling multi-stage tasks~\cite{zheng2023ddcot,li2024mvbench}. To overcome these challenges, we build Long-Horizon Planning and Reasoning in VLN (LHPR-VLN) based on the NavGen platform. \textit{LHPR-VLN} is the first LH-VLN benchmark that consists of 3,260 tasks with an average of 150 task steps.
This large-scale benchmark captures the depth and variety required for evaluating long-horizon VLN, encompassing a wide range of sub-task structures and navigation complexities.
Additionally, traditional coarse-grained success rates (SR) are inadequate for complex tasks, as task complexity makes it difficult for overall success rates to accurately reflect model capabilities. 
Therefore, we propose three new metrics for more thorough evaluation: Conditional Success Rate (CSR), CSR weighted by Ground Truth (CGT), and Independent Success Rate (ISR), to assess success for each subtasks, capturing the model’s performance at each step and offering a more detailed evaluation of execution across the full scope of LH-VLN challenges.

% : Conditional Success Rate (CSR), CSR weighted by Ground Truth (CGT), and Independent Success Rate (ISR), to assess success for each subtasks, capturing the model’s performance at each step and offering a more detailed evaluation of execution across different stages. 

% To further verify a model’s ability to accurately understand and execute instructions across long, multi-stage task sequences, we designed a dynamically granular benchmark, Dynamic LH-VLN \note{跟前面Navgen那一段的关系是什么，是否可以合在一起讲？}. This benchmark modifies subtask requirements or the current environment state during navigation, testing whether the model truly understands language instructions and navigates based on them rather than relying on path inertia. This refined evaluation approach not only reveals the model’s adaptability and dependence on instructions in complex scenarios but also provides a new perspective and effective metric for future LH-VLN performance improvements.

%% 需要方法图确定了再改
Existing VLN methods typically rely on discretizing the environment into static points for path prediction, limiting adaptability in complex, dynamic settings~\cite{etpnav_2024, zhou2025navgpt,wu2024embodied,long2024instructnav}. 
To bridge this gap and enhance real-world applicability in LH-VLN tasks, we introduce a Multi-Granularity Dynamic Memory (MGDM) module to enhance the model's adaptability and memory handling. 
The MGDM module operates by integrating both short-term and long-term memory mechanisms. 
While short-term memory blurring and forgetting functions help the model focus on recent, relevant information, long-term memory retrieval pulls in key historical data from previous navigation steps~\cite{song2023llm}.
This combination allows the model to adjust to environmental changes and retain context over extended sequences, addressing the challenges of sustained reasoning and adaptive re-planning in dynamic environments. 
With MGDM, we achieve state-of-the-art performance on the LH-VLN task, demonstrating its effectiveness in maintaining coherent decision-making and robust navigation over long, multi-stage tasks. Our contributions can be summarized as follows:

% we designed a Multi-Granularity Dynamic Memory (MGDM) module based on the CoT feedback. MGDM incorporates adaptive short-term memory blurring and forgetting mechanisms, alongside a long-term memory retrieval system. By integrating relevant examples from the dataset (long-term memory) with historical information from the current task (short-term memory), MGDM enables the model to adjust to environmental changes. 

%To balance memory demands with computational efficiency, we introduce an adaptive blurring strategy. Specifically, when short-term memory reaches a set capacity threshold, the model gradually attenuates non-essential information, retaining only critical content rather than completely discarding the oldest memory. This flexible memory management approach reduces computational load while enhancing model efficiency and performance in complex LH-VLN tasks. 

\begin{itemize}
    \item We propose the LH-VLN task, a new task designed to evaluate agents in complex, multi-stage navigation tasks requiring sustained reasoning and adaptability.
    \item We develop NavGen, an automated data generation platform that produces a high-quality, long-horizon dataset, enabling scalable task diversity and improved data utility.
    \item We introduce the LHPR-VLN benchmark with 3,260 tasks, each averaging 150 steps, and propose three new metrics for detailed, sub-task-level evaluation.
    \item We present the MGDM model, designed to enhance model adaptability in dynamic settings through combined short-term and long-term memory mechanisms.
\end{itemize}

% \begin{itemize}
%     \item To mitigate the lack of high-quality data and improve data utilization efficiency for LH-VLN, we propose an automated data generation platform, named NavGen, which constructs datasets with complex task structures through a bidirectional, multi-granularity generation approach.
%     \item To comprehensively evaluate the long-horizon vision-language navigation task, we propose the first large-scale benchmark Long-Horizon Planning and Reasoning in VLN (LHPR-VLN) and introduce three new metrics. 
%     \item To achieve robust LH-VLN,  we propose a novel Multi-Granularity Dynamic Memory (MGDM) module based on the CoT feedback, which integrates short-term memory blurring with long-term memory retrieval to enable flexible navigation in dynamic environments.
% \end{itemize}

% 简要介绍当前具身导航发展
% 说明当前具身导航面临问题：
% 高质量数据的缺少以及现有benchmark老旧
% 在更加通用、复杂场景表现不佳
% 过于依赖大模型性能，以及基于waypoint predictor的方法带来不够”真实“的性能
% 一些更加具体的说明可以参考这篇论文：Navigating the Nuances: A Fine-grained Evaluation of Vision-Language Navigation
% 我们要设计更加通用、面向需求的框架
% 设计更加复杂、贴近现实需求的任务，探索任务间因果关系对模型的影响
% 类似 RoboGen，可以持续生成数据
% 记忆的设计是当前复杂任务模型设计的重点，思维链引入，可解释性

%% file: sec/2_related_work.tex
\section{Related Work}
\label{sec:related_work}

\begin{figure*}[!t]
  \centering
    \includegraphics[width=0.85\textwidth]{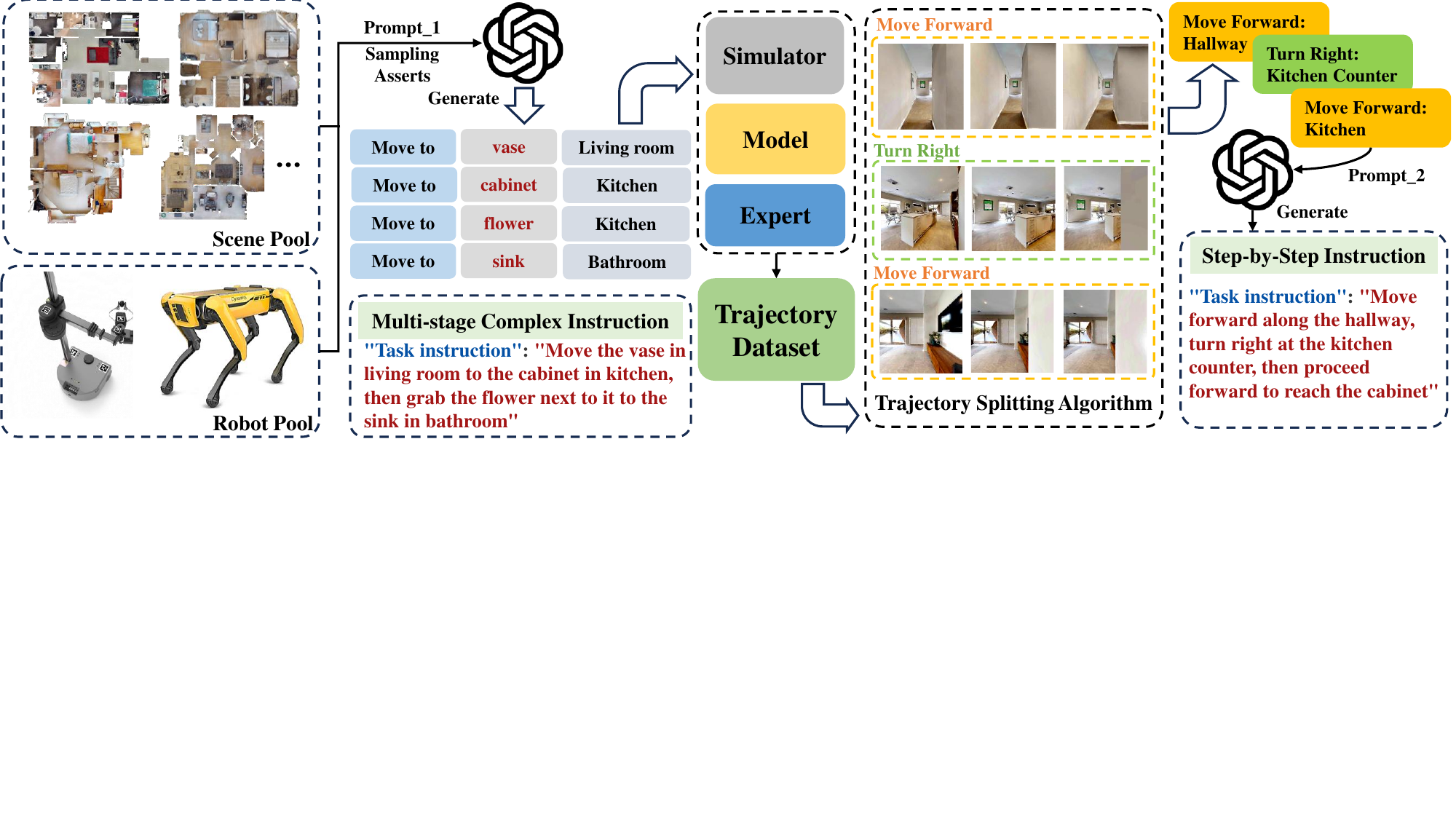}
    \vspace{-10pt}
    \caption{The NavGen data generation platform. The forward generation generates LH-VLN complex tasks and corresponding subtasks by prompting GPT-4 with sampling asserts. The sampled assets are deployed on the simulator. Based on the navigation model or expert decisions, corresponding trajectory data is generated. In the backward generation, the trajectory of each subtask is split into action-label pairs by trajectory splitting algorithm according to the trajectory type, these pairs are then input into GPT-4 to generate step-by-step tasks.
    %NavGen generates multi-stage long-range navigation tasks based on scene assets and robot configurations, and customizes deployment models and experts in the simulator to generate target trajectories. Based on the target trajectories, our trajectory segmentation algorithm divides the trajectories into action-related blocks. These blocks are annotated by an image annotation model to form action-annotation pairs. The action-annotation pairs then generate step-by-step VLN tasks. NavGen realizes data generation from assets to instructions, to trajectories, and back to instructions, greatly enhancing the richness and usability of the data.
    }
      \vspace{-10pt}
    \label{fig:navgen}
\end{figure*}
\subsection{Vision-Language Navigation}
Embodied Vision-Language Navigation (VLN) aims to enable agents to perform navigation tasks in complex environments based on language instructions. Current methods advance in three main directions: map-based strategies, waypoint prediction, graph-based approaches, and large-model predictions. Map-based strategies, such as VLN-VER~\cite{liu2024volumetric} and HNR-VLN~\cite{wang2024lookahead}, employ volumetric representations or neural radiance fields to facilitate spatial understanding and exploration by the agent. Modular designs like those in FILM~\cite{minfilm} integrate language instructions with environmental perception, enhancing task efficiency. The second category, waypoint prediction-based methods, includes models such as ETPNav~\cite{an2024etpnav} and MultiPLY~\cite{hong2024multiply}, which optimize navigation through key-point prediction and environmental graph learning, thereby supporting improved generalization across discrete and continuous environments \cite{hong2022bridging}. Additionally, large language model-based approaches, including NaviLLM~\cite{navillm_2024} and NaViD~\cite{zhang2024navid}, excel at interpreting complex instructions by tightly integrating language reasoning with visual tasks. However, existing methods often remain limited to single-stage tasks and lack consistent planning for long-horizon, multi-stage tasks. %To address this limitation, we propose the Long-Horizon Vision-Language Navigation (LH-VLN) task along with the data generation platform, large-scale evaluation benchmark, and dedicated method, emphasizing consistent planning and decision-making across sequential tasks to provide a more practical solution for realistic navigation challenges.

\subsection{Benchmark for Vision-Language Navigation}
The progression of VLN tasks has been propelled by a range of datasets, each introducing unique challenges and enhancing evaluation benchmarks for embodied agents performing tasks in complex environments. Early datasets, such as Room-to-Room (R2R)~\cite{R2R_2018} and its extension Room-for-Room (R4R)~\cite{jain2019stay}, focus on step-by-step navigation through predefined paths with fine-grained instructions based on static images, while later datasets like VLN-CE~\cite{vln_ce_2020} shift towards continuous navigation in dynamic spaces, requiring more flexible decision-making. More recent datasets, including CVDN~\cite{thomason2020vision}, REVERIE~\cite{REVERIE_2020}, and SOON~\cite{SOON_2021}, further broaden the scope of VLN by integrating dialogue history, object localization, and complex instruction comprehension, pushing agents to understand high-level natural language commands and locate specific targets. Meanwhile, OVMM~\cite{OVMM_2023} and Behavior-1K~\cite{BEHAVIOR-1K_2023} add layers of complexity by incorporating navigation, manipulation, and object interaction, simulating extended real-world tasks that involve multiple sub-tasks. IVLN~\cite{ivln} and Goat-Bench~\cite{khanna2024goatbench}  allow the agent to continuously complete multiple independent single-target navigation tasks while maintaining memory. Despite these progresses, there is still a notable gap in benchmarks that support LH-VLN with multi-stage sub-tasks in highly complex environments. %To address this, we introduce a comprehensive solution for the LH-VLN task, encompassing both dataset construction and baseline reference models.

%% file: sec/3_muti_stage_causal_navigation_task.tex
\section{Platform, Benchmark, and Metrics}
\label{sec:multi-stage_causal_navigation_task}
We developed a data generation platform named NavGen, specifically designed to support the data needs of the LH-VLN task. Based on this platform, we created the LHPR-VLN benchmark to evaluate model performance in terms of long-term planning capabilities within this task.

\subsection{NavGen} % : Data Generation while Navigation}
The NavGen platform integrates automated data generation with a bi-directional generation mechanism to produce task instructions and associated trajectory data. The two-pronged approach includes forward data generation, which focuses on complex LH-VLN task creation, and backward data generation, which decomposes multi-stage navigation sub-tasks into granular, actionable steps, shown in Fig. \ref{fig:navgen}.

\subsubsection{Forward Data Generation}
In the forward data generation phase, we utilize GPT-4 to create task instructions by synthesizing scene assets and robot configurations, as shown in Fig.~\ref{fig:navgen}. Specifically, our scene assets come from the HM3D dataset~\cite{yadav2023habitat}, which offers a rich collection of 3D panoramic scenes annotated semantically across 216 settings, providing an extensive foundation for task creation. Additionally, robot configurations are carefully tailored to different robotic platforms, such as Boston Dynamics' Spot and Hello Robot's Stretch, each with unique camera heights, task spaces, and sensor parameters to accommodate a variety of tasks. 
% 29.16 1131.8
\begin{table*}\scriptsize  \setlength{\tabcolsep}{3pt}
  \centering
  \begin{tabular}{l|cccccc}
    \toprule
    Benchmark & Avg. Instruction Length & Avg. Task Steps & Simulator & Task Type & Scenes & Task Num \\
    \midrule
        R2R \cite{R2R_2018} & 29 & $<$8 & Matterport3D & Step-by-step Nav & 90 & 21567 \\
    REVERIE \cite{REVERIE_2020}& 18 & $<$8 & Matterport3D & Obj Loco-nav & 90 & 21702 \\
    VLN-CE \cite{vln_ce_2020} & 30 & 55.88 & Habitat & Step-by-step Nav & 90 & 4475 \\
    FAO \cite{SOON_2021} & 39 & 10 & Matterport3D & Obj Loco-nav & 90 & 3848 \\
    % IVLN \cite{ivln} & - & - & M3D\&Habitat & Iterative VLN & 72 & 789\\
    Behavior-1k \cite{BEHAVIOR-1K_2023} & 3.27 & - & OmniGibson & Complex Housework & 50 & 1000 \\
    IVLN \cite{ivln} & - & - & M3D\&Habitat & Iterative VLN & 72 & 789\\
    Goat-Bench \cite{khanna2024goatbench} & - & - & Habitat & Iterative VLN & 181 & 725360\\
    LHPR-VLN (Ours) &  18.17 & 150.95 & Habitat & Multi-stage VLN & 216 & 3260 \\
    \bottomrule
  \end{tabular}
  \vspace{-10pt}
  \caption{Comparison to VLN benchmarks.}
    \vspace{-10pt}
  \label{tab:benchmark}
\end{table*}
% , and task goals $G$
With these assets and configurations as the initial resource pool, a custom-designed prompt serves as the input for GPT-4, which combines scene details $S$ and robot configurations $R$. Then GPT-4 outputs an instruction list $D_{ins}=\mathcal{G}(S,R,\text{prompt}_{1})$, including the sub-task and multi-stage instructions, and $\mathcal{G}$ is denoted the GPT-4. This list is imported into the Habitat3 simulator $Sim$, where an expert model or a well-trained navigation model guides the agent $A$ through the task, which the expert model is a navmesh model and greedy pathfinder algorithm built from Habitat~\cite{gupta2017cognitive, kumar2018visual}. The simulator autonomously generates trajectories $D_{traj}$, the foundational data for subsequent splitting into task segments:

\begin{equation}
    D_{traj} = Sim(D_{ins}, S, A, \mathbf{OR}(M, E))
\end{equation}
where $\mathbf{OR}$ represents that either $M$ or $E$ can be used.

\subsubsection{Backward Data Generation}

After obtaining the trajectory through forward task generation, we decompose the trajectory of complex tasks and create step-by-step VLN tasks for each trajectory segment. The trajectory decomposition algorithm (more detail can be found in the supplementary material) splits complex task trajectories into multiple single-stage navigation task trajectories. Within a single-stage navigation goal trajectory, the algorithm divides the trajectory into segments representing ``move forward,” ``turn left,” ``turn right,” and ``bypass forward.” By using a dynamic sliding window, the algorithm continuously searches for all the longest continuous action segments within the trajectory. These continuous action segments serve as the basic units of action instructions in step-by-step navigation tasks. For each segment, the RAM image annotation model~\cite{zhang2024recognize} provides high-confidence visual annotations. These annotations, coupled with action instructions, are input as prompts into GPT-4 to generate VLN tasks for step-by-step guidance, thereby creating a refined set of decomposed single-stage navigation tasks.

% The backward generation phase splits the generated complex trajectories into fine-grained, stepwise tasks. Each sub-task within a complex trajectory undergoes segmentation through the Trajectory Splitting Algorithm (Algorithm \ref{algorithm:Traj_Split}), which organizes the task into basic action primitives like ``move forward,” ``turn left,” ``turn right,” and ``bypass forward.” This decomposition uses a dynamic sliding window to identify continuous action segments, maintaining coherence within each sub-task. For each segment, the RAM image annotation model provides high-confidence visual annotations. These annotations, coupled with action instructions, are processed to generate descriptive VLN steps, thereby creating a refined set of decomposed single-stage navigation tasks.

% 通用的 MSCN 任务生成框架，训练模型同时源源不断生成数据，模块化，可迁移

\subsection{The LHPR-VLN Benchmark}
% 首先讲当前bencmark有哪些不足：
% 任务过短，复杂度不够高，不足以对应当前大模型发展；
% 综合场景多样性、任务多样性不足；
% 指令形式不切合现实（人类无法给出如此细致准确指令）；
% 在接受具有内在因果关系的任务时，模型会有怎样的表现？是否能够真正理解指令？这是否能够促进具身智能领域的可解释性研究？

% 随后提出任务设置

% 最后与其他benchmark比较

Our LHPR-VLN benchmark defines a complex task that includes multiple single-stage subtasks. For an LHPR-VLN task, the basic format is: ``Find something somewhere, and take it to something somewhere, then…". Each complex task involves locating an object at a specified initial location and transporting it to a designated target location, potentially encompassing two to four sequential navigation sub-tasks. The embodied agent needs to sequentially complete these single-stage navigation tasks to ultimately fulfill the instruction. For each single-stage navigation task, the agent must approach within a 1-meter geodesic distance of the target object, ensuring the object is positioned within a 60-degree horizontal field of view to maintain task fidelity.

%In our proposed LHPR-VLN dataset, we develop a multi-stage task structure that requires the agent to execute a sequence of single-stage navigation sub-tasks. Each complex task involves locating an object at a specified initial location and transporting it to a designated target location, potentially encompassing two to four sequential navigation sub-tasks. The embodied agent needs to sequentially complete these single-stage navigation tasks to ultimately fulfill the instruction. For each single-stage navigation task, the agent must approach within a 1-meter geodesic distance of the target object, ensuring the object is positioned within a 60-degree horizontal field of view to maintain task fidelity.

\begin{figure}[!t]
    \centering
    \includegraphics[width=0.85\linewidth]{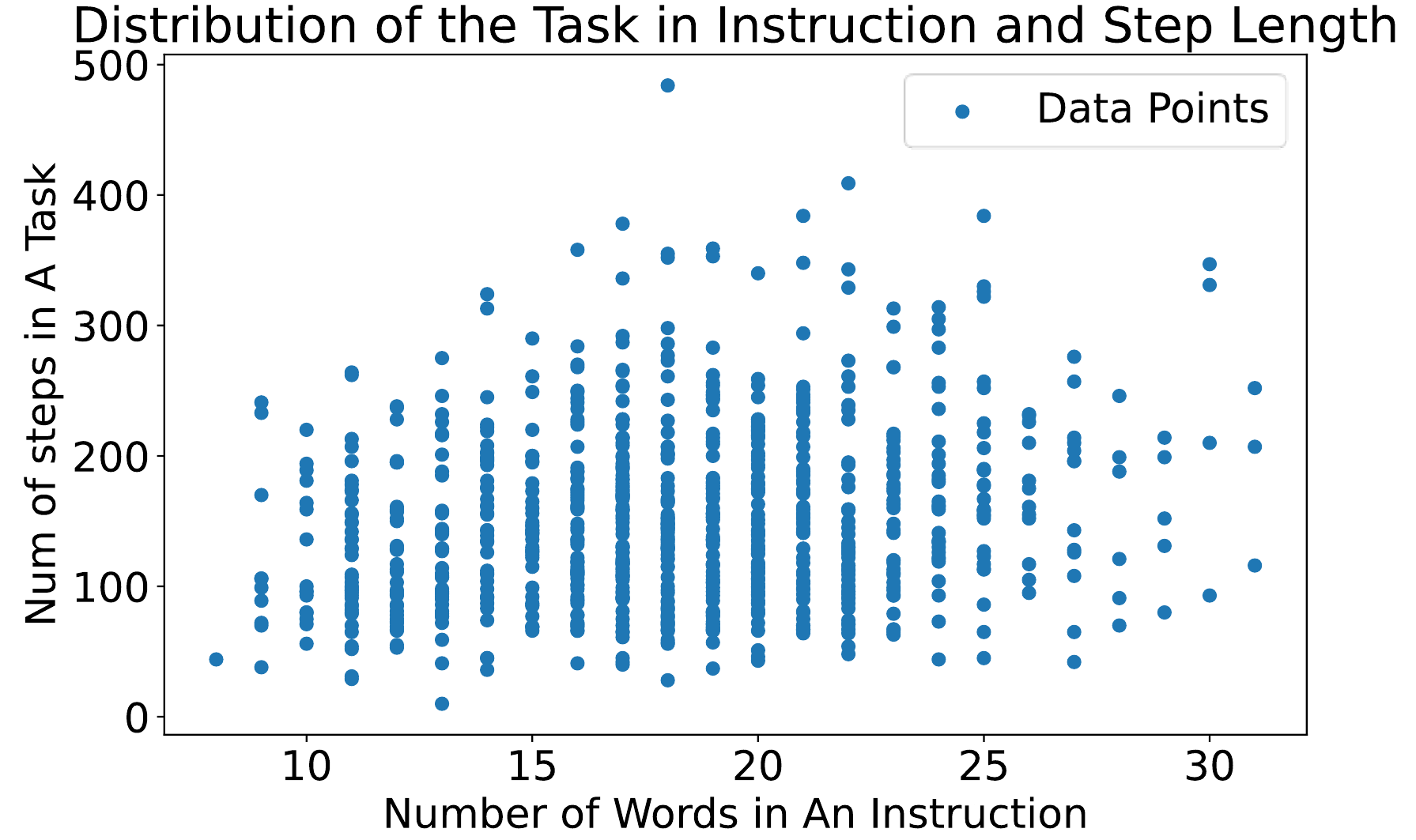}
    \vspace{-10pt}
    \caption{Overview of the  LHPR-VLN benchmark statistics. In our statistics, Spot and Stretch robot-type tasks account for 50.5\% and 49.5\%, respectively. LH-VLN tasks containing 2, 3, and 4 subtasks account for 39.0\%, 52.4\%, and 8.6\%, respectively.}
        \vspace{-10pt}
    \label{fig:enter-label}
\end{figure}
% (a) Distribution of the Robot. (b) Distribution of the task length. (c) Distribution of instruction and step length of the task. We consider 2, 3, 4 subtasks as Short, Medium, and Long Task, respectively.
Throughout navigation, the agent acquires observational data from three perspectives ($+60\degree$, $0\degree$, $-60\degree$) and is permitted to execute fundamental actions: turn left, move forward, turn right, and stop. When the agent selects the ``stop" action, the sub-task is deemed complete, and task success is evaluated based on the agent’s final positional state relative to the target. Table \ref{tab:benchmark} presents a comparison between representative VLN
benchmarks, our LHPR-VLN is the first LH-VLN benchmark, containing 3,260 multi-stage and step-by-step VLN tasks from 216 complex scenes, with an average of 150 task action steps and 18.17 instruction length. 

\begin{figure*}[!t]
    \centering
    \includegraphics[width=0.9\linewidth]{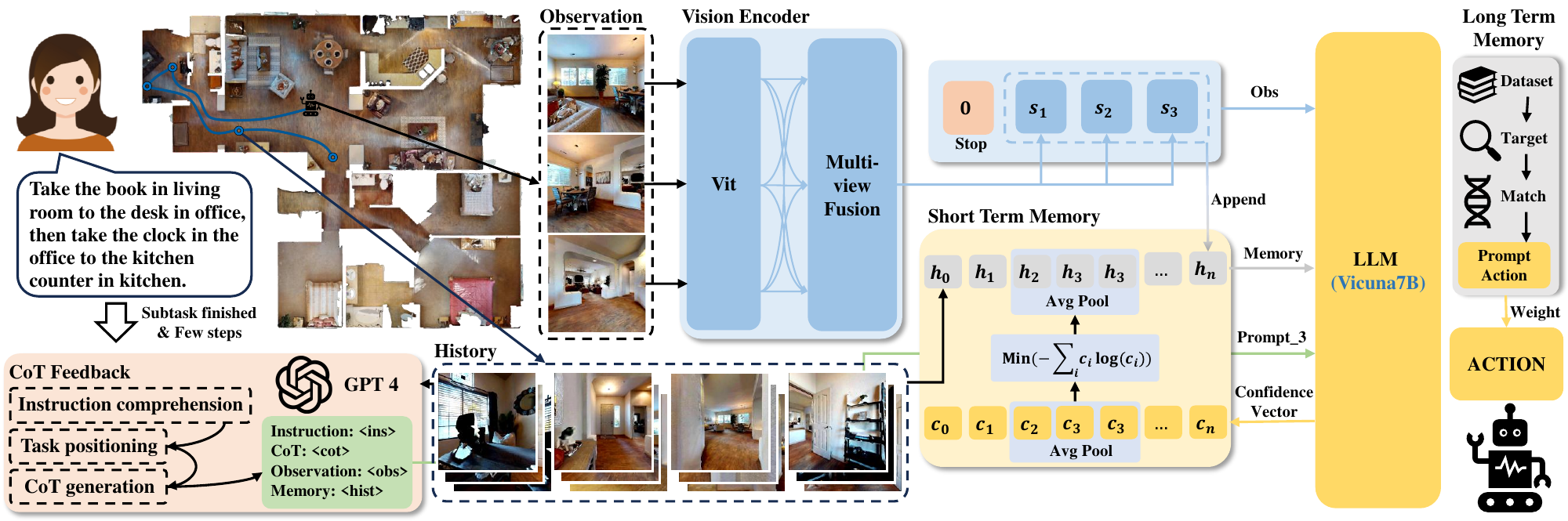}
    \vspace{-10pt}
    \caption{The framework of the Multi-Granularity Dynamic Memory (MGDM) model. The CoT feedback module receives task instructions and, based on historical observation of corresponding memory, generates a chain of thought and constructs language prompts. The short-term memory module aims to minimize the entropy of the confidence vector, using pooling operations to forget and blur the memory sequence. The long-term memory module selects and matches data from the dataset to weight the decisions of the LLM, ultimately determining the action to be executed by the agent.
    %When a sub-task is completed or after a certain number of steps, the agent will perform CoT feedback. Based on instructions and a sequence of history observations, the agent will carry out instruction comprehension, positioning, and CoT generation to adjust and reflect on the current task progress. When short-term memory reaches a certain length, it minimizes the entropy of the confidence vector for corresponding actions generated by the LLM. The observation representation, short-term memory, and prompts are input to the LLM to obtain action decisions. Long-term memory identifies the most relevant action prompts from the dataset, which are then weighted in the decision-making process to produce the final action.
    }
    \vspace{-10pt}
    \label{fig:method}
\end{figure*}

\subsection{Reasonable Metrics}
To rigorously assess model performance in the LH-VLN task, we introduced specialized metrics, complementing the standard evaluation metrics (Success Rate (SR), Oracle Success Rate (OSR), Success weighted by Path Length (SPL), and Navigation Error (NE)). These new metrics include Independent Success Rate (ISR), Conditional Success Rate (CSR), and CSR weighted by Ground Truth (CGT). ISR quantifies the success rate of each sub-task individually, providing insight into independent sub-task completion rates. CSR evaluates the success of the overall complex task, as the outcome of each sub-task impacts the subsequent ones, thus encapsulating interdependencies in the task sequence. 
\begin{equation}
    ISR = \sum_{j=0}^M \sum_{i=0}^N \frac{ s_{j,i}}{M \cdot N}
\end{equation}
where $M$ is the numble of tasks, and $N$ is the number of sub-tasks in $\textrm{Task}_j$.
The CSR metric is calculated as follows:
\begin{equation}
    CSR = \sum_{j=0}^M \sum_{i=0}^N \frac{s_{j,i}(1+(N-1)s_{i-1})}{M \cdot N^2}
\end{equation}
where $s_{j,i}$ denotes the success of the $i$-th sub-task in $\textrm{Task}_j$. 

CGT further refines CSR by incorporating ground truth weighting, to account for deviations in path difficulty. CGT is calculated as:
\begin{equation}
    CGT = \sum_{j=0}^M \sum_{i=0}^N \frac{P_i}{P} \cdot \frac{s_{j,i}(1+(N-1)s_{j,i-1})}{M \cdot N}
\end{equation}

We also designed a metric Target Approach Rate (TAR) based on NE to reflect the model's performance in cases where the navigation success rate is relatively low. The relevant settings can be found in the supplementary materials.

Furthermore, the multi-granularity task instructions generated by the NavGen platform allow us to test the model’s responsiveness to various instruction types within the same trajectory. This testing approach not only facilitates an analysis of the agent’s focus during navigation but also enables a robust evaluation of task comprehension and execution across complex scenarios through these novel metrics. Thus, these new metrics provide a comprehensive evaluation of model performance in LH-VLN tasks.

%% file: sec/4_method.tex
\section{Multi-Granularity Dynamic Memory Model} % : Navigate with CoT and Memory} 
% Baselin需要重新取一个名字

To achieve robust LH-VLN, our Multi-Granularity Dynamic Memory (MGDM) model follows the general VLN pipeline and comprises three essential components: the base model, the Chain-of-Thought (CoT) Feedback module, and Adaptive Memory Integration and Update (AMIU), as shown in Fig. \ref{fig:method}. These components enable robust performance in LH-VLN, addressing challenges related to spatial awareness \cite{luo2025dspnet}, instruction comprehension \cite{chen2025cross}, and task continuity \cite{EXPRESSBench} across long-horizon sequences.

\label{sec:method}
\subsection{Base Model}
% 基础模型结构。图像编码，自然语言编码，融合编码，输入大模型，输出动作选择
The base model aligns with the standard structure of VLN models. For scene observation, the model encodes multi-directional visual information using a pre-trained visual encoder vit. Each observed image $I_i$ is processed into visual features $v_i$. To integrate scene information across multiple directions, a Transformer encoder is used for multi-view feature fusion. The directional image features $\{v_i\}_{i=1}^n$ are processed through the Transformer encoder, resulting in contextually enriched representations $\{o_i\}_{i=1}^n$ that capture inter-relational information across views.

Each directional view is distinguished by embedding directional tokens (`left,' `front,' `right') to construct a comprehensive scene representation $S$: 
\begin{equation} 
S = [\mathcal{E}(\text{`left'}), o_{\text{left}}, ..., \mathcal{E}(\text{`right'}), o_{\text{right}}] \end{equation} 
where $\mathcal{E}$ denotes the embedding layer. For historical observations $H_i$, each previous scene is encoded similarly, with stepwise embeddings added to capture temporal relations, establishing sequential order within the observation history:
\begin{equation} 
H_{n+1} = [\mathcal{E}(1), h_1, ..., \mathcal{E}(n), h_n] 
\end{equation} 
The scene and historical representations are then combined into a unified prompt, which is fed into the large language model (LLM) $\mathcal{G}$ to select the next action:
\begin{equation} 
a_{n+1} = \mathcal{G}(\mathcal{E}(\text{prompt}_{3}), S, H_n) 
\end{equation}
  
\subsection{Navigate with CoT and Memory}
% 行动前任务分解以及思维链生成模块
% 我觉得这里需要结合记忆来写，应该说是CoM不是CoT，然后下一章是如何增强CoM的M
To address the limited interpretability and susceptibility to ``hallucinations"~\cite{varma2024ravl} in LLM-based VLN models (wherein the agent completes tasks without true comprehension), we introduce a Chain-of-Thought (CoT) \cite{wei2022chain} Feedback module that receives task instructions and, based on historical observation of corresponding memory, generates a chain of thought and constructs language prompts.  This module aims to enhance the agent's reasoning capability by iteratively refining its task understanding and action planning.

\textbf{CoT Feedback.} At the beginning of each sub-task and periodically during navigation, the CoT Feedback module receives task instructions, current observation, and history visual observations in memory, along with the prompt, are input into GPT-4 to generate the chain of thought $\textrm{CoT}=\textrm{GPT-4}(\textrm{Obs, Hist, Instruction, Prompt})$. GPT-4 uses past observations and task instructions to establish the current task context, which implies comprehensive task understanding. The task is then decomposed based on this understanding, guiding the agent’s immediate actions. This reflective process enables the agent to adjust and refine its interpretations, improving task comprehension and execution.

% 记忆模块：短期记忆与长期记忆
% 记忆模糊
\textbf{Adaptive Memory Integration and Update.} 
Previous VLN works often used visual encoding from past observations as memory, which is typically effective. However, in LH-VLN tasks, the lengthy task duration causes an excessive accumulation of memory, making this approach impractical. Moreover, existing methods often discard the oldest memories to maintain a fixed-length memory sequence or just discard some memories that the model thinks inappropriate~\cite{etpnav_2024}, which inadvertently removes critical information. To mitigate these limitations, we design an Adaptive Memory Integration and Update (AMIU) module incorporating short-term memory, long-term memory, and a memory blurring and forgetting process.

Short-term memory $M_{st}$ is structured from historical observation encoding, capturing temporally ordered observations as the agent moves through the environment:
\begin{equation}
    M_{st} = \{h_i\}_{i=0}^n
\end{equation}
When the memory length $n$ reaches a set maximum $N$, dynamic forgetting is triggered. Each memory element $h_i$ has an associated confidence score $c_i=\mathcal{G}(\cdot)_i$, representing the model's confidence in corresponding action. The memory sequence $M_{st}$ thus has an associated confidence vector $C=\{c_i\}_{i=o}^n$.

The forgetting module employs a ``pooling function" that we define it as $\mathcal{P}$. $\mathcal{P}(C)_i$ represents the pooling operation with a window size of 2 applied to the $i_{th}$ element and its neighboring elements in $C$, which reduces its length by one:
\begin{equation}
    \mathcal{P}(C)_i=\{c_1,...,\textrm{AvgPool}(c_{i-1},c_i,c_{i+1}),...,c_n\}=C_i\\
\end{equation}
where $C_i \in \mathbb{R}^{n-1}$. We apply the pooling operation to each element in $C$ separately, obtaining $\{C_i\}_{i=0}^n=\{\mathcal{P}(C)_i\}_{i=0}^n$. We then calculate the entropy of each $C_i$ and identify the pooling index with the smallest entropy:
\begin{equation}
    \text{arg}\min_i(-\sum_{j=1}^{n-1}s_j\log s_j), s_j = \frac{C_{i,j}}{\sum_{j=0}^{n-1}C_{i,j}}
\end{equation}

The same pooling operation is applied to the $M_{st}$ elements corresponding to the pooling index and add new short-term memory to maintain the memory sequence.
\begin{equation}
    M_{st}=\mathcal{P}(M_{st})_i + h_n^*
\end{equation}
% The forgetting module employs a ``pooling function" that selectively blurs adjacent memory elements. It minimizes entropy across the confidence vector, retaining more confident memories while merging less reliable entries:
% \begin{equation}
%     \textrm{min}(-\sum_{i=1}^ns_i\log s_i), s_i = \frac{\text{AvgPool}(C)_i}{\sum_{j=0}^n\text{AvgPool}(C)_j}
% \end{equation}

% Selected confidence elements are pooled with neighboring values, and pooled memories are appended to the short-term memory sequence:
% \begin{equation}
%     ST=\{h_1,...,\textrm{AvgPool}(h_{c-1},h_c,h_{c+1}),...,h_n\} + h_n^*
% \end{equation}

Long-term memory $M_{lt}$ serves as a reinforcement mechanism. As the agent navigates, long-term memory retrieves relevant observations and actions based on target $T$ from the dataset, matching them with the agent's current observation to provide guidance. The retrieval process selects the top $k$ matching observation-action pairs, which are weighted to inform the current decision vector. This memory is sourced from the LHPR-VLN dataset, reinforcing prior learning:
\begin{equation}
    M_{lt} = \textrm{Dataset}(T)=\{obs_j, act_j\}^m_{j=1}
\end{equation}
Thus, the indices of the selected $M_{lt}$ can be formulated as:
\begin{equation}
    I_{k}=\text{argsort}_{t=0}^{k}(\{\frac{obs_j \cdot v}{\sqrt{\sum_{i=1}^{n_v}obs_{j,i}^2} \cdot \sqrt{\sum_{i=1}^{n_v}v_i^2}}\}^m_{j=1})
\end{equation}
The action decision $a$ is weighted by averaging the retrieved actions:
\begin{equation}
    a = a \cdot \text{avg}(\{act_t\}_{t=0}^{k})
\end{equation}
where $a$ is the current decision vector. The final cross-entropy loss is computed between the model's decision $a$ and the expert's decision $e$ at current action:
\begin{equation}
\text{arg}\min_{\Theta}\mathcal{L}(a,e) = \text{arg}\min_{\Theta} (-\sum_{i=0}^n a_i\log(e_i))
\end{equation}

%% file: sec/5_experiment.tex
\section{Experiment}
\label{sec:experiment}

\subsection{Experimental Settings}

%\subsubsection{Simulation Settings}
% habitat 3
% sensor
\noindent \textbf{Simulator:} We conduct experiments in Habitat3~\cite{gupta2017cognitive, kumar2018visual}, which provides a continuous 3D scene platform for VLN. Additionally, we perform experiments in Isaac Sim, which has high-quality scene rendering and physical interactions.

\noindent \textbf{Sensors:} For each action step, the agent receives RGB observations from there directions of front, left (+60°), and right (-60°). Depth images for these three directions can also be customized.

\begin{table*}
  \centering\scriptsize
  \setlength{\tabcolsep}{10pt}
  \begin{tabular}{l|c|ccccc|ccccc}
    \toprule
    \multirow{2}{*}{Method} & \multirow{2}{*}{Type} & \multicolumn{5}{c|}{2-3 Subtasks} & \multicolumn{5}{c}{3-4 Subtasks} \\

    && SR$\uparrow$ &  NE$\downarrow$ & ISR$\uparrow$ & CSR$\uparrow$ & CGT$\uparrow$ &SR$\uparrow$  & NE$\downarrow$ & ISR$\uparrow$ & CSR$\uparrow$ & CGT$\uparrow$ \\
    \midrule
    % L3MVN \cite{L3MVN_2023} & Zero-shot \\
    % InstructNav \cite{instructnav_2024}& Zero-shot \\
    Random & - & 0. &14.09 & 0. & 0. & 0. & 0. & 10.91 & 0. & 0. & 0.\\
    % ETPNav \cite{etpnav_2024} & Trained\\
    GLM-4v prompt \cite{glm2024chatglm} & Zero-shot  & 0. & 15.63 & 0. & 0. & 0.& 0.  & 10.97 & 0. & 0. & 0.\\
    NaviLLM \cite{navillm_2024} & Pretrain & 0. & 12.11 & 0. & 0. & 0.& 0.  & 10.04 & 0. & 0. & 0.\\
    NaviLLM \cite{navillm_2024} & Finetuned & 0. & 12.24 & 0. & 0. & 0.& 0.  & 9.79 & 3.54 & 2.53 & 5.24\\
    GPT-4 + NaviLLM & Pretrain & 0. & 12.23 & 0. & 0. & 0.& 0.  & 10.00 & 4.37 & 2.91 & 5.23\\
    \textbf{MGDM (Ours)} & Finetuned & 0. & \textbf{3.54} & 0. & 0. & 0.& 0.  & \textbf{1.23} & \textbf{4.69} & \textbf{3.30} & \textbf{5.83}\\
    \bottomrule
  \end{tabular}
  \vspace{-10pt}
  \caption{Performance comparison in LH-VLN Task with different task length.}
    \vspace{-10pt}
  \label{tab:result_1}
\end{table*}

\noindent \textbf{Actions:} We provide atomic actions for the agent, including `move forward' (+0.25), `turn left' (+30°), `turn right' (-30°), and `stop'. When the agent performs the stop action, the current task (or sub-task) is considered complete. We also provide a coordinate-based movement option.

\noindent \textbf{Scene Assets:} Our scene assets are primarily from HM3D~\cite{yadav2023habitat}, which includes 216 large-scale indoor 3D reconstructed scenes with semantic annotations. Besides, we use HSSD~\cite{10657917hssd}, which includes 211 high-quality indoor scenes, to test the data generation with NavGen. 

\noindent \textbf{Robot Configurations:} The robots include the Stretch robot from Hello Robot and the Spot robot from Boston Dynamics. Stretch has a wheeled base and a manipulator with a structural frame, while Spot is a quadruped robot dog capable of mounting a mechanical arm on its back.

%\subsubsection{Model Training Settings}
\noindent \textbf{Training Settings:} We alternately use imitation learning and trajectory-based supervised learning. 
The LLM is Vicuna 7B v0~\cite{vicuna2023}, and the visual encoder is the ViT model from EVA-CLIP-02-Large~\cite{sun2023eva}. The visual encoder remains frozen during training. 
In the training phase, we utilize the Adam optimizer with a learning rate of 3e-5.

%\subsubsection{Metrics}
\noindent \textbf{Metrics:} Besides our metrics \textbf{ISR}, \textbf{CSR} , and \textbf{CGT}, we also used traditional metrics~\cite{R2R_2018}, including \textbf{SR} (Success Rate), \textbf{SPL} (Success weighted by Path Length), \textbf{OSR} (Oracle Success Rate), and \textbf{NE} (Navigation Error). For SR, OSR and SPL, the task is considered successful only when all sub-tasks in a LH-VLN task are completed correctly in the logical sequence of instructions. For NE, only when the agent takes the action of `stop', the NE counts.

% \begin{figure}[!t]
%     \centering
% \includegraphics[width=0.95\linewidth]{sec/img/5-traj.pdf}
%     \vspace{-10pt}
%     \caption{Tasks and trajectories generated with NavGen in Habitat and Isaac Sim.}
%         \vspace{-10pt}
%     \label{fig:Data-Generation}
% \end{figure}
\vspace{-5pt}
\subsection{Baseline Models}
\begin{itemize}
    % \item L3MVN \cite{L3MVN_2023} utilizes zero-shot learning to comprehend long-distance multi-modal navigation tasks.
    % \item InstructNav \cite{instructnav_2024}: 
    \item ETPNav \cite{etpnav_2024}: ETPNav is a graph-based navigation model where the agent’s current and historical observations are modeled as graph nodes. %A waypoint predictor identifies navigable waypoints, and based on the graph and candidate nodes, the nav module predicts the next action node and updates the graph, ultimately finding the target.
    \item GLM-4v prompt \cite{glm2024chatglm}: GLM-4v is a state-of-the-art vision-language model. To evaluate the performance of vision-language models in LH-VLN tasks, we use prompt engineering to guide GLM-4v to produce reasonable outputs and test its actual performance.
    \item NaviLLM \cite{navillm_2024}: NaviLLM is the state-of-the-art model for navigation in discrete environments. %It uses visual fusion encoding, memory, and language embedding to directly predict the next action through a large language model. 
    We adapted this approach to continuous environments and fine-tuned it on the dataset to evaluate its performance in LH-VLN.
    \item GPT-4 + NaviLLM: To evaluate the performance of traditional single-stage models in LH-VLN with the assistance of a LLM to decompose complex tasks, we combined GPT-4 with NaviLLM. GPT-4 first decomposes the complex task into several sub-task, and NaviLLM then executes each sub-task sequentially.
\end{itemize}

\begin{figure*}[!t]
    \centering
\includegraphics[width=0.92\linewidth]{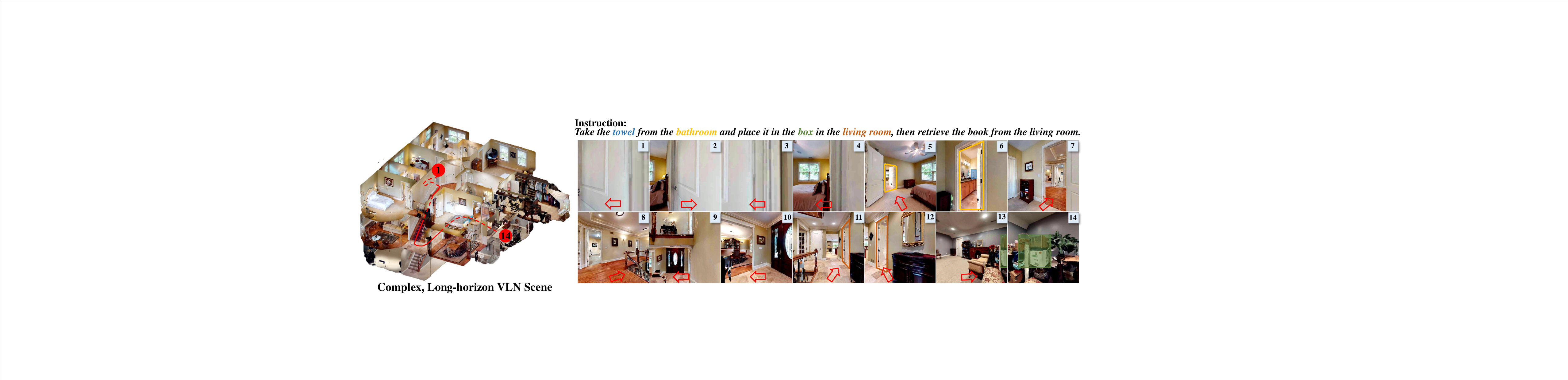}
    \vspace{-10pt}
    \caption{Visualization of a partially successful long-horizon navigation of our MGDM. We highlight aligned landmarks by colored bounding boxes in images and words in the instruction using the same color. In the first navigation segment, the agent looks for a towel in the bathroom. It successfully finds both the bathroom and the towel but does not enter the bathroom or gets close enough to the towel for the task to be marked as successful. In the next phase, the agent successfully finds the box in the living room.}
        \vspace{-10pt}
    \label{fig:visual}
\end{figure*}

% \subsection{Data Generation}
% We conduct experiments for data generation with NavGen in Habitat and Isaac Sim. Based on NavGen, we use HM3D scene assets and Habitat3's built-in navigation algorithm to generate tasks and trajectories in Habitat3, and HSSD scene assets with a D* Lite-based path planning algorithm to generate trajectories and tasks in Isaac Sim. Tasks and trajectories shown in the Figure \ref{fig:Data-Generation}, and more detailed experimental results are provided in the appendix.

% \subsubsection{Metrics}
% Success Rate(\textbf{SR}), Oracle Success Rate(\textbf{OSR}),SR penalized by Path Length(\textbf{SPL}) ,Navigation Error(\textbf{NE})

% Independent Success Rate(\textbf{ISR}): The average success rate of independent subtasks

% Conditional Success Rate(\textbf{CSR}): The success rate of conditional execution tasks:
% \begin{equation}
%   CSR = \sum_{i=0}^N \frac{s_i(1+(N-1)s_{i-1})}{N^2}
%   \label{eq:CSR}
% \end{equation}

% CSR weighted by Path Length(\textbf{CGT}): The success rate of conditional execution tasks weighted by path length:
% \begin{equation}
%   CGT = \sum_{i=0}^N \frac{P_i}{P} \cdot \frac{s_i(1+(N-1)s_{i-1})}{N}
%   \label{eq:CGT}
% \end{equation}
\subsection{Result Analysis}
We test baseline models on LH-VLN task and its corresponding step-by-step trajectories with LHPR-VLN benchmark. Through these tests, we aim to answer the following questions: \textbf{Q1:} Can existing models understand and complete multi-stage complex tasks with limited information? \textbf{Q2:} How to understand the relations between multi-stage complex tasks and single-stage simple tasks? \textbf{Q3:} What is the significance of memory in multi-stage complex tasks?

\textbf{RQ1:} For ETPNav, due to the inherent limitations of its waypoint predictor, even with only three viewpoint RGBD settings, the model still fails to effectively predict navigable points, despite being designed to handle invalid navigation points and deadlock states.  The performance of each model in LH-VLN task is shown in Table \ref{tab:result_1}. As seen, all models perform poorly. In the relatively short LH-VLN tasks with 2-3 subtasks, the SR, ISR, CSR, and CGT of all models are 0. This indicates that these models are unable to complete even a single subtask. In the longer LH-VLN tasks with 3-4 subtasks, only fine-tuned NaviLLM, GPT-4+NaviLLM, and our MGDM can complete some subtasks. This suggests that existing models cannot effectively understand and complete multi-stage complex tasks with limited information.

\textbf{RQ2:} To explore the relation between multi-stage complex tasks and single-stage simple tasks, we test the combination of the single-stage navigation model NaviLLM with GPT-4 task decomposition. By using GPT-4 to decompose complex tasks, NaviLLM can sequentially perform several single-object navigation tasks. In Table \ref{tab:result_1}, it can be seen that the performance of GPT-4+NaviLLM shows some improvement compared to the pre-trained NaviLLM and fine-tuned NaviLLM, especially in ISR, where it improves by 23\% compared to the fine-tuned NaviLLM. This indicates a significant performance improvement on individual subtasks, highlighting its single-stage navigation ability.

\begin{table}\scriptsize
  \centering
    \setlength{\tabcolsep}{10pt}
  \begin{tabular}{l|ccccc}
    \toprule
    Method  & SR$\uparrow$ & OSR$\uparrow$ & SPL$\uparrow$ & NE$\downarrow$  \\
    \midrule
    % L3MVN \cite{L3MVN_2023}  \\
    % InstructNav \cite{instructnav_2024} \\
    Random & 0. & 0 & 0. & 8.59\\
    % ETPNav \cite{etpnav_2024} \\
    GLM-4v prompt \cite{glm2024chatglm}& 0. & 11.1 & 0. & 6.50\\
    NaviLLM \cite{navillm_2024} & 6.67 & 6.67 & 2.86 & 10.17\\
    \textbf{MGDM (Ours)} & 0. & \textbf{26.92} & 0. & \textbf{1.70} \\
    \bottomrule
  \end{tabular}
    \vspace{-10pt}
  \caption{Performance comparison in step-by-step LH-VLN task.}
    \vspace{-20pt}
  \label{tab:result_2}
\end{table}

However, the performance of the GPT-4+NaviLLM method is still slightly lower than that of our MGDM, which has been specifically designed for complex tasks, especially in CGT. In fact, the CGT metric for GPT-4+NaviLLM is even lower than that of fine-tuned NaviLLM. Since CGT is weighted based on the length of the ground truth, this result suggests that our MGDM is better at completing longer and more difficult subtasks. The reason may be that our MGDM directly executes complex tasks can maintain more coherent and complete memories, which help it accomplish more complex tasks. Additionally, the advantage in CSR further indicates that MGDM has a better comprehensive understanding of multi-stage LH-VLN tasks. 

Actually, combining task decomposition for complex tasks with single-stage navigation models can improve the performance of single-stage models on complex tasks to some extent. However, this approach also leads to a lack of holistic understanding of complex tasks, as well as incomplete and fragmented memory.

\textbf{RQ3:} Furthermore, all models perform better in ISR, CSR, and CGT on LH-VLN tasks with 3-4 subtasks than on those with 2-3 subtasks. This may be due to the fact that while longer and multi-stage tasks may be more difficult, the memory accumulated from previous stages can help the VLN model complete subtasks in subsequent stages. This may suggest the significance of developing VLN models for multi-stage complex tasks. The tendency of navigation target distribution in the LH-VLN task with different numbers of subtasks and task settings may also influence this result. Relevant details can be found in the supplementary materials. It is worth noting that our MGDM has a relatively low NE. when tasks are so difficult that the model performs poorly, NE reflects the gap between the model performance and success. This suggests that our MGDM may have greater potential for LH-VLN. Additionally, in the step-by-step tasks shown in Table \ref{tab:result_2}, although our MGDM has higher OSR and lower NE, its SR and SPL metrics are both 0. This indicates that our MGDM faces an issue in effectively determining whether the goal has been achieved.
% \begin{table*}
%   \centering
%   \begin{tabular}{l|ccccccc|ccccccc}
%     \toprule
%     \multirow{2}{*}{Method} & \multicolumn{7}{c|}{Spot} & \multicolumn{7}{c}{Stretch} \\

%     & SR & OSR & SPL & NE & ISR & CSR & CGT &SR & OSR & SPL & NE & ISR & CSR & CGT \\
%     \midrule
%     L3MVN \cite{L3MVN_2023} &  \\
%     InstructNav \cite{instructnav_2024}&  \\
%     ETPNav \cite{etpnav_2024} & \\
%     GPT-4 + EPTNav & \\
%     HNR \cite{HNR_2024} & \\
%     GLM-4 prompt \cite{glm2024chatglm} & \\
%     NaviLLM \cite{navillm_2024} & \\
%     Ours & \\
%     \bottomrule
%   \end{tabular}
%   \caption{Performance of SOTA models in LH-VLN Task with different view.}
%   \label{tab:result_2}
% \end{table*}

\subsection{Ablation Studies}
\begin{table}
    \setlength{\tabcolsep}{10pt}
  \centering\scriptsize
  \begin{tabular}{l|cccc}
    \toprule
    Method & NE$\downarrow$ & ISR$\uparrow$ & CSR$\uparrow$ & CGT$\uparrow$  \\
    \midrule
    % L3MVN \cite{L3MVN_2023}  \\
    % InstructNav \cite{instructnav_2024} \\
    MGDM w/o Adap Mem & 4.44 & 0. & 0. & 0.\\
    MGDM w/o LT Mem & 11.13 & 2.20 & 1.27 & 2.08\\
    % ETPNav \cite{etpnav_2024} \\
    MGDM w/o CoT & 2.45 & 0. & 0. & 0.\\
    \textbf{MGDM} & \textbf{1.23} & \textbf{4.69} & \textbf{3.30} & \textbf{5.83}\\
    \bottomrule
  \end{tabular}
    \vspace{-10pt}
  \caption{Ablation results.}
    \vspace{-20pt}
  \label{tab:result_3}
\end{table}
We performed ablation studies on multi-granularity dynamic memory module, long term memory module and the chain-of-thought (CoT) feedback module, with results shown in Table \ref{tab:result_3}. As observed, the model's performance is significantly affected whether the CoT feedback module, long term memory module or the multi-granularity dynamic memory module is ablated. This indicates the crucial role of chain-of-thought generation and memory in the model's ability to solve LH-VLN tasks. From the perspective of NE, especially the multi-granularity dynamic memory module, it has significant impact on model's performance. This is also reflected in the visualization analysis of a successful long-horizon navigation example (see Figure \ref{fig:visual}). The agent's actions are very chaotic at the beginning (1-3 steps). It only acts effectively once the memory sequence reaches a certain length. This further underscores the importance of memory module design for LH-VLN tasks.

%% file: sec/6_conclusion.tex
%\vspace{-5pt}
\section{Conclusion}
\label{sec:conclusion}

We address the challenges of long-horizon vision-language navigation (LH-VLN) from three aspects: platform, benchmark, and method. Specifically, we develop an automated data generation platform NavGen, which constructs datasets with complex task structures 
and improves data utility.
We also construct the LHPR-VLN benchmark, which provides three new  metrics for detailed, sub-task-level evaluation. Additionally, we present the MGDM model, designed to enhance
model adaptability in dynamic settings through combined 
short-term and long-term memory mechanisms, achieving outstanding performance on the LH-VLN task.
%Our contributions establish a unified framework for LH-VLN research, making significant strides toward bridging the gap between simulation and real-world LH-VLN navigation applications. 

%% file: sec/X_suppl.tex
\clearpage
\setcounter{page}{1}
\maketitlesupplementary

\section{Symbol Table}
\label{sec:Symbol Table}
\vspace{-10pt}
\begin{table}[!h]
  \centering\small% \scriptsize
  \begin{tabular}{c|l}
    \toprule
    \textbf{Symbol} & \textbf{Explanation}  \\
    \midrule
    % L3MVN \cite{L3MVN_2023}  \\
    % InstructNav \cite{instructnav_2024} \\
    $I_i$ & Observed image from \textbf{view} $i$ \\
    % ETPNav \cite{etpnav_2024} \\
    $v_i$ & Visual features of \textbf{view} $i$\\
    $o_i$ & Fusion visual features of \textbf{view} $i$ \\
    $S$ & Scene representation of \textbf{current observation} \\
    $\mathcal{E}$ & Tokenizer \\
    $ h_i$ & History representation of \textbf{step} $i$\\
    $ H_{n+1} $ & The memory set of the previous $n$ steps obtained \\
    & at the $n+1_{th}$ step\\
    $ \mathcal{G}$ & Large language model\\
    $a_n$ & Model decision action at \textbf{step} $n$\\
    $e_n$ & Expert decision action at \textbf{step} $n$\\
    $n_v$ & The number of viewpoints in the observation.\\
    $I_{k}$ & The indices of the $k$ selected elements\\
    $M_{st}$ & Short term memory\\
    $M_{lt}$    & Long term memory\\
    $C$ &   Confidence vector generated from $\mathcal{G}$\\
    $\mathcal{P}$   & Pooling function\\
    \bottomrule
  \end{tabular}
    \vspace{-7pt}
  \caption{Symbol Table}
    \vspace{-15pt}
  \label{tab:Symbol_Table}
\end{table}

\section{Benchmark}
\label{sec:benchmark}
\subsection{Trajectory Splitting Algorithm}
We design a Trajectory Splitting Algorithm \ref{algorithm:Traj_Split} for NavGen to backward-generate Step-by-Step Navigation Task from Navigation Trajectory.

\begin{algorithm}[!h]
\caption{Trajectory Splitting Algorithm }
\KwIn{Navigation trajectory $D_{traj}$, Image annotation model $Ram$}
\KwOut{Trajectory segments $Seg$}
$actions=[turn\ left, turn\ right], s=[]$;\\
\For{action in actions}{
    $i=0$;\\
    \While{$i < D_{traj}.length-3$}{
        $window = D_{traj}[i:i+3]$;\\
        \eIf{$window.count(action) \geq 2$}{
            $indices = window.index(action)$;\\
            $s.append((index[0], index[-1], action))$;\\
            $i = index[-1]+1$;\\
        }{$i=i+1$}
    }
}
$s.sort()$;\\
$merge\_s=[], c\_start, c\_end, c\_label = s[0]$;\\
\For{$s_i\ in\ s[1:]$}{
    $start, end, label = s_i$;\\
    \eIf{$start \leq c\_end + 3$ and $label == c\_label$}{
        $c\_end = max(c\_end, end)$;\\
    }{
        $merge\_s.append((c\_start, c\_end, c\_label))$;\\
        $c\_start, c\_end, c\_label = s_i$;\\
    }
}
$merge\_s.append((c\_start, c\_end, c\_label))$;\\
$Seg = [], last\_end=-1$;\\
\For{$strat, end, act\ in\ merge\_s$}{
    \If{$last\_end+2 < start$}{
        $Seg.append((move\ forward, Ram(D_{traj}[last\_end + 1, start - 1])$;\\
        }
    $Seg.append((act, Ram(D_{traj}[start - 1, end + 1])$;\\
    $last\_end = end$;\\
}

\label{algorithm:Traj_Split}
\end{algorithm}
\begin{figure*}[!t]
    \centering
\includegraphics[width=1\linewidth]{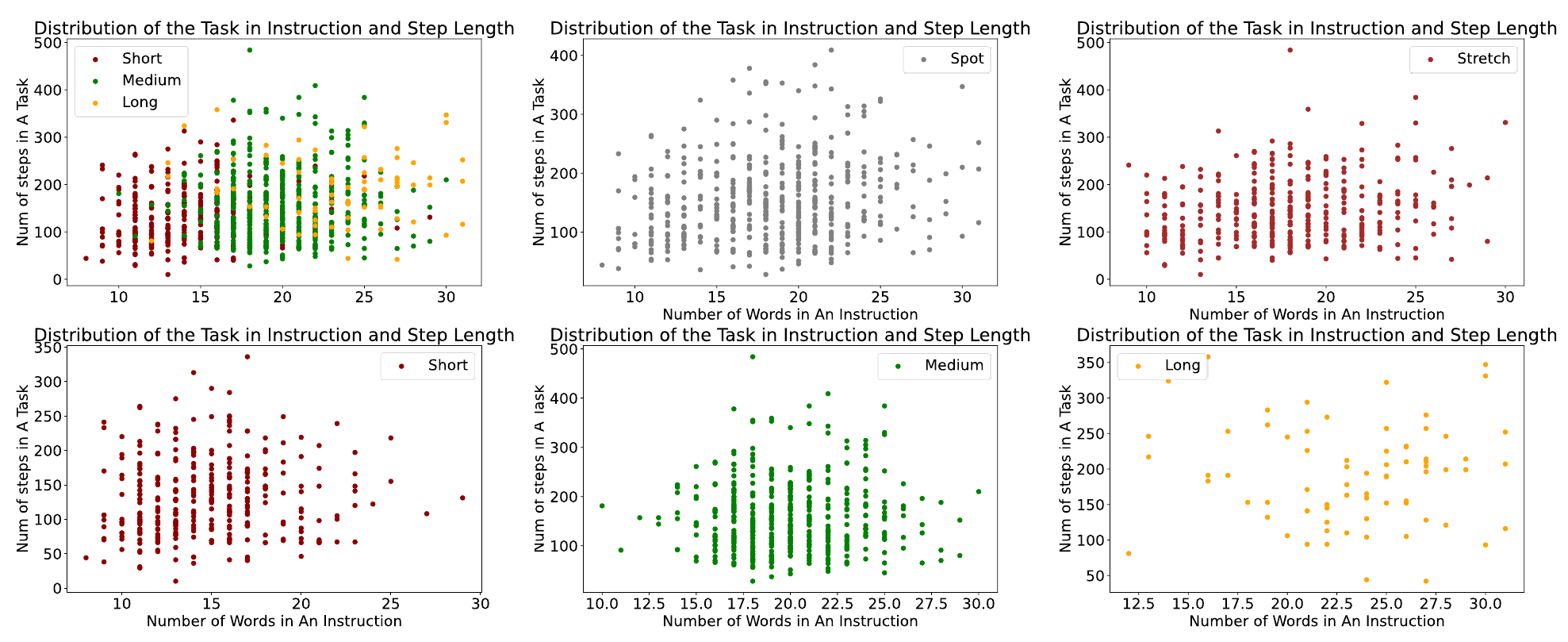}
\vspace{-20pt}
    \caption{Statistic of the LH-VLN dataset distribution based on task length and robot configuration. We consider
2, 3, 4 subtasks as Short, Medium, and Long Task, respectively.}
\vspace{-7pt}
    \label{fig:statics}
\end{figure*}
\subsection{Dataset Statistics}
We conducted statistical analysis based on the complex task training set of the LH-VLN dataset, and the results are shown in Figure \ref{fig:statics}.

We also conducted statistical analysis on more detailed data regarding task goals at different stages in the dataset, as shown in Table. \ref{tab:benchmark}.
\begin{table*}[t]\scriptsize
  \centering
    \setlength{\tabcolsep}{10pt}
  \begin{tabular}{l|cccccc}
    \toprule
    Type  & Task Steps & Nav Dis & Obj Num & Floor Span & Reentry Rate & Area Association \\
    \midrule
    % L3MVN \cite{L3MVN_2023}  \\
    % InstructNav \cite{instructnav_2024} \\
    2 subtasks & 68.39 & 11.60, 11.23, -, - & 288.43 & 1.05 &27.44\%  & 1.02\\
    % ETPNav \cite{etpnav_2024} \\
    3 subtasks & 53.30 & 10.32, 11.47, 4.74, -&357.05&1.44&42.11\% & 1.89\\
    4 subtasks & 51.23 & 9.75, 10.45, 4.14, 9.39&273.37&1.88&48.84\% & 2.12\\
    % spot \\
    % stretch \\
    \bottomrule
  \end{tabular}
    \vspace{-7pt}
  \caption{Detailed dataset statistics. \emph{Task Steps} represents the average steps for each subtask. \emph{Nav Dis} refers to the average geodesic distance between the agent and the target at the start of each stage, \emph{Obj Num} is the average number of objects in the scene, \emph{Floor Span} represents the average number of floors the task spans, \emph{Reentry Rate} is the ratio of tasks where the agent needs to retrace its steps, indicating that the agent may have previously observed the target of the current subtask in prior tasks, and \emph{Area Association} refers to the average number of identical subtask areas.
  }
    \vspace{-14pt}
  \label{tab:benchmark}
\end{table*}

\subsection{Extra Metric}
We designed a metric Target Approach Rate (TAR) based on NE to reflect the model’s performance in cases where the navigation success rate is relatively low. 
For the $i$-th subtask of the $j$-th task, we calculate the \textbf{Target Approach Rate (TAR)}: 
\begin{equation}
tar_{j,i}=1-\frac{max(NE_{j,i}-D_s,\ 0)}{max(NE_{j,i}, GT_{j,i})}
\end{equation}
where $NE_{j,i}$ is NE of the $i$-th subtask of the $j$-th task, $D_s$ is the distance to be considered success, $GT_{j,i}$ is ground truth of the $i$-th subtask of the $j$-th task.

\subsection{Cases Study}
\begin{figure*}[!t]
    \centering
\includegraphics[width=0.95\linewidth]{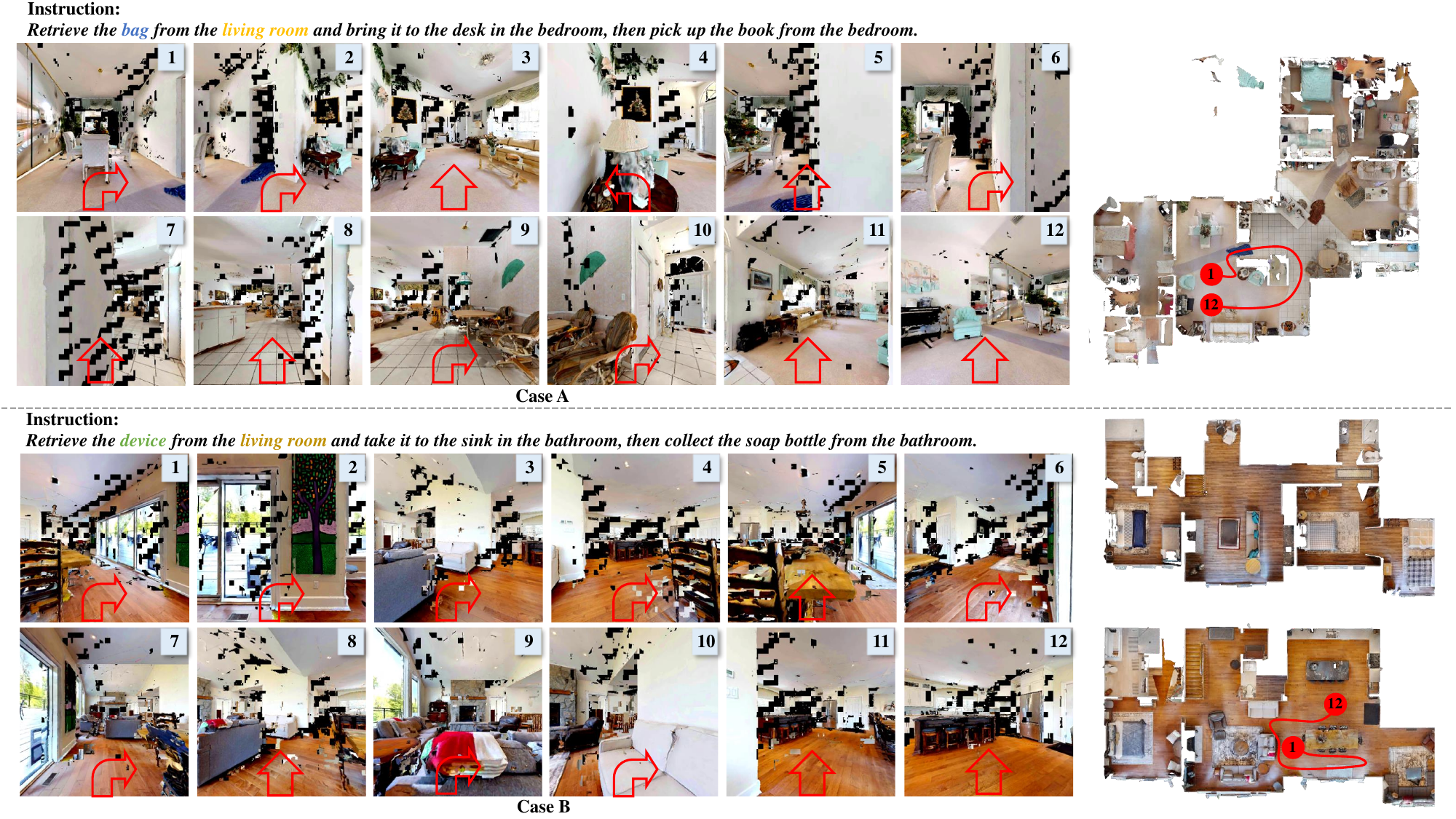}
\vspace{-10pt}
    \caption{Part of trajectories of two Task.}
    \vspace{-10pt}
    \label{fig:case}
\end{figure*}

In the Figure \ref{fig:case}, we present two cases for analysis.

For Case A, the agent quickly found the living room and identified the item on the table in Observation 3 as the target. However, in Observation 4, it determined that it was not a "bag", prompting a strategy change and further exploration of the scene (Observations 5–12). After completing the scene exploration, the agent resumed the search for the target.

For Case B, the agent was initially unable to determine the current scene and decided to rotate in place (Observations 1–4). It then tentatively searched for the living room. Upon realizing the current direction did not lead to the living room, it turned around (Observations 6–7) and moved toward the living room area (Observations 8–10). Since it did not find the "device" in the living room, the agent chose to explore another direction (Observations 11–12).

\section{Extra Experiments}

\subsection{Prompt Used}
We present our NavGen forward task generation prompt ($prompt_1$ \ref{fig:prompt1}), backward task generation prompt ($prompt_2$ \ref{fig:prompt2}), and MGDM model prompt ($prompt_3$ \ref{fig:prompt3}). $Prompt_3$ is designed with reference to ~\cite{navillm_2024}.

\subsection{Data Generation}
\begin{figure*}[!h]
    \centering
\includegraphics[width=0.95\linewidth]{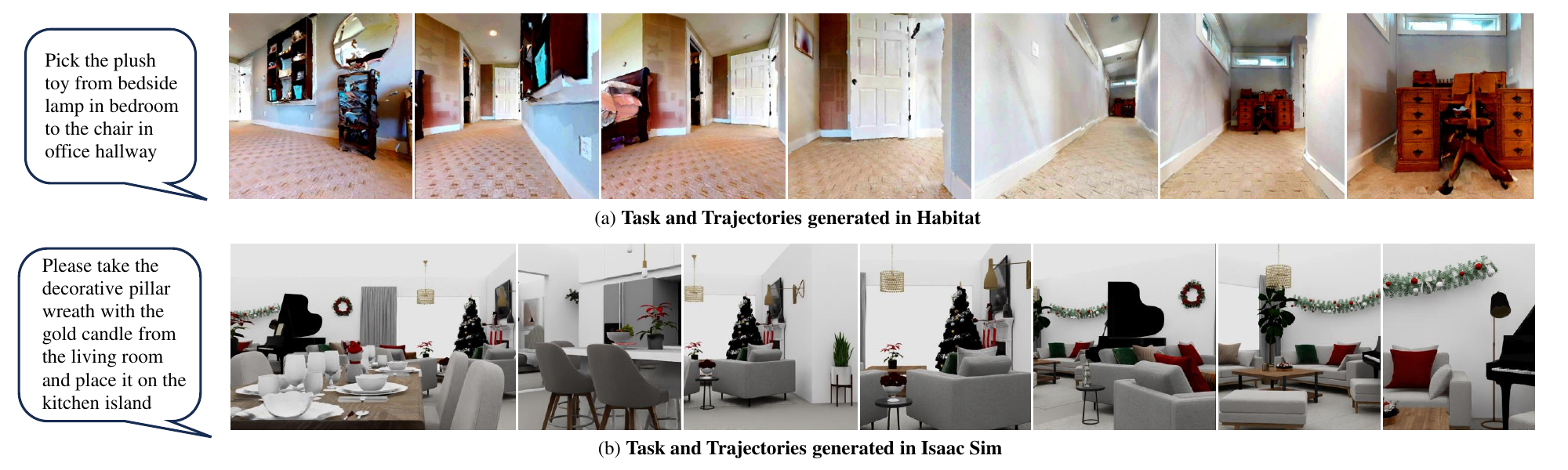}
\vspace{-10pt}
    \caption{Tasks and trajectories generated with NavGen in Habitat and Isaac Sim.}
    \vspace{-10pt}
    \label{fig:Data-Generation}
\end{figure*}

We conduct experiments for data generation with NavGen in Habitat and Isaac Sim. Based on NavGen, we use HM3D scene assets and Habitat3's built-in greedy pathfinder algorithm to generate tasks and trajectories in Habitat3, and HSSD scene assets with a D* Lite-based~\cite{Koenig2002Dlite} path planning algorithm to generate trajectories and tasks in Isaac Sim. Tasks and trajectories shown in the Figure \ref{fig:Data-Generation}.

\subsection{More details on the experimental setup}
Specifically, we designed two training approaches. One involves alternating between supervised learning and imitation learning at each step during the training of the main section. The other is the two-stage~\cite{navillm_2024} training approach: pre-training with supervised learning for the first stage, followed by further training using reinforcement learning.

For supervised learning, we train on the trajectory dataset from the LH-VLN dataset. The agent obtains observations, location information, and action labels from the dataset without being directly deployed in the simulator. In imitation learning, the agent is deployed in the simulator, where it acts based on task instructions within the simulated environment. Expert actions are provided by Habitat 3's greedy pathfinder algorithm, and the agent learns to mimic these expert actions.

For additional experiments, we tested the state-of-the-art zero-shot model, InstructNav~\cite{instructnav_2024}. InstructNav decomposes instructions into dynamic chain-of-navigation using GPT-4. Based on DCoN, InstructNav optimizes navigation strategies from various perspectives using a simple numerical map and ultimately generates action decisions. Compared to other models, InstructNav utilizes additional information, including depth maps of the current viewpoint, top-down maps of the scene, and partial use of Habitat 3's greedy pathfinding interface.

\begin{table*}[!t]
  \centering
  \setlength{\tabcolsep}{8pt}
  \begin{tabular}{l|ccccc|ccccc}
    \toprule
    \multirow{2}{*}{Method} & \multicolumn{5}{c|}{Spot} & \multicolumn{5}{c}{Stretch} \\

    & SR$\uparrow$ &  NE$\downarrow$ & ISR$\uparrow$ & CSR$\uparrow$ & CGT$\uparrow$ &SR$\uparrow$  & NE$\downarrow$ & ISR$\uparrow$ & CSR$\uparrow$ & CGT$\uparrow$ \\
    \midrule
    Random  & 0. & 14.97 & 0. & 0. & 0.& 0.  & 10.48 & 0. & 0. & 0.\\
    InstructNav \cite{instructnav_2024}  & 0. & 11.48 & 0. & 0. & 0.& 0.  & 7.70 & 0. & 0. & 0.\\
    GLM-4v prompt \cite{glm2024chatglm}  & 0. & 15.97 & 0. & 0. & 0.& 0.  & 10.55 & 0. & 0. & 0.\\
    NaviLLM (Pretrain) \cite{navillm_2024} & 0. & 10.27 & 0. & 0. & 0.& 0.  & 11.43 & 0. & 0. & 0.\\
    NaviLLM \cite{navillm_2024}  & 0. & 10.97 & 2.19 & 1.31 & 2.72 & 0.  & - & 0 & 0 & 0\\
    GPT-4 + NaviLLM  & 0. & 10.15 & \textbf{3.85} & \textbf{2.00} & \textbf{3.96} & 0.  & 11.59 & 2.33 & 1.59 & 2.47\\
    \textbf{MGDM (Ours)} & 0. & \textbf{3.44} & 3.73 & 1.92 & 3.81 & 0.  & \textbf{1.37} & \textbf{2.82} & \textbf{2.06} & \textbf{3.25}\\
    \bottomrule
  \end{tabular}
  \vspace{-7pt}
  \caption{Performance of SOTA models in LH-VLN Task under different Robot configurations.}
  \vspace{-5pt}
  \label{tab:result_robot}
\end{table*}

\begin{table*}[!t]
  \centering\scriptsize
  \setlength{\tabcolsep}{10pt}
  \begin{tabular}{l|c|ccccc|ccccc}
    \toprule
    \multirow{2}{*}{Method} & \multirow{2}{*}{Type} & \multicolumn{5}{c|}{2-3 Subtasks} & \multicolumn{5}{c}{3-4 Subtasks} \\

    && SR$\uparrow$ &  NE$\downarrow$ & ISR$\uparrow$ & CSR$\uparrow$ & CGT$\uparrow$ &SR$\uparrow$  & NE$\downarrow$ & ISR$\uparrow$ & CSR$\uparrow$ & CGT$\uparrow$ \\
    \midrule
    % L3MVN \cite{L3MVN_2023} & Zero-shot \\
    % InstructNav \cite{instructnav_2024}& Zero-shot \\
    % ETPNav \cite{etpnav_2024} & Trained\\
    Random & - & 0. &14.09 & 0. & 0. & 0. & 0. & 10.91 & 0. & 0. & 0.\\
    % ETPNav \cite{etpnav_2024} & Trained\\
    InstructNav \cite{instructnav_2024} & Zero-shot  & 0. & 10.80 & 0. & 0. & 0.& 0.  & 8.38 & 0. & 0. & 0.\\
    GLM-4v prompt \cite{glm2024chatglm} & Zero-shot  & 0. & 15.63 & 0. & 0. & 0.& 0.  & 10.97 & 0. & 0. & 0.\\
    NaviLLM \cite{navillm_2024} & Pretrain & 0. & 12.11 & 0. & 0. & 0.& 0.  & 10.04 & 0. & 0. & 0.\\
    NaviLLM \cite{navillm_2024} & Finetuned & 0. & 12.24 & 0. & 0. & 0.& 0.  & 9.79 & 3.54 & 2.53 & 5.24\\
    GPT-4 + NaviLLM & Pretrain & 0. & 12.23 & 0. & 0. & 0.& 0.  & 10.00 & 4.37 & 2.91 & 5.23\\
    MGDM (Llama3/Two Stage) & Finetuned & 0. & 13.20 & 0. & 0. & 0.& 0.  & 10.02 & 4.60 & 3.07 & 5.38\\
    MGDM (Vicuna/Alternate) & Finetuned & 0. & \textbf{3.54} & 0. & 0. & 0.& 0.  & \textbf{1.23} & 4.69 & 3.30 & 5.83\\
    MGDM (Vicuna/Two Stage) & Finetuned & 0. & 10.44 & 0. & 0. & 0.& 0.  & 8.78 & \textbf{5.13} & \textbf{3.42} & \textbf{6.00}\\
    \bottomrule
  \end{tabular}
  \vspace{-7pt}
  \caption{Performance comparison in LH-VLN Task with different task length. We add InstructNav, MGDM (Llama3/Two Stage), MGDM (Vicuna/Two Stage) for r comparison. (Vicuna/Two Stage) stands for LLM and training set used.}
  \vspace{-7pt}
  \label{tab:result_add}
\end{table*}

Furthermore, we replaced the large language model in MGDM to test its compatibility with different large models and the impact of these models on test results, following the two-stage training approach. The model used for replacement testing is Llama 3.1 8B Instruct~\cite{llama}. Due to GPU memory constraints, we froze the first five layers of the model's language layers.

\subsection{More experimental results}
First, we supplement the test results under different robot configurations (Spot, Stretch) in Table \ref{tab:result_robot}.

Based on the Random results, the difficulty of tasks executed by the Spot and Stretch robots differs. The NE (Navigation Error) for Spot is greater than that for Stretch, indicating that the agent, on average, is farther from the objects in tasks involving the Spot robot. Under this premise, the data in the table \ref{tab:result_robot} shows that, apart from the zero-shot model InstructNav and GLM-4v prompt, which perform relatively evenly across both types of tasks, other models generally perform better on Spot robot tasks compared to Stretch robot tasks. Given that the LH-VLN training set has a relatively balanced distribution of both types of tasks, and the NaviLLM pretrain model (which was not trained on the LH-VLN dataset) exhibits the same trend, we believe the performance difference may stem from the lower viewpoint of the Spot robot. This configuration excludes the influence of smaller objects, making it easier for the agent to approach the target area.

Additionally, in tasks with the Spot robot configuration, MGDM's performance is slightly lower than NaviLLM+GPT-4. Notably, for tests involving NaviLLM-related models, their performance shows a significant advantage under the Spot robot configuration. This suggests that the NaviLLM model is relatively better suited for Spot-related tasks. This advantage may stem from NaviLLM's pretraining data being more closely aligned with the Spot robot configuration, giving it a certain edge in this aspect. In contrast, MGDM, due to the design of its memory module, is more likely to "remember" small objects in the scene. This may reduce the performance gap between the Spot and Stretch robot configurations, leading to more balanced results across both setups.

Furthermore, we include InstructNav from the additional experiments, as well as MGDM models employing different large language models and training strategies, for comparison. As shown in Table \ref{tab:result_add}, due to its Dynamic Chain-of-Navigation module's ability to comprehend and decompose complex tasks, InstructNav performs exceptionally well among zero-shot models. However, some of InstructNav's configurations, such as trajectory value maps that avoid historical paths, are not well-suited for multi-stage complex tasks.

The impact of replacing Llama 3 is limited, which may be attributed to the performance differences between Vicuna 1.5 and Llama 3, as well as the effects of varying training configurations on the model. MGDM trained with the two-stage setup achieved better performance on ISR, CSR, and CGT metrics compared to alternating training but performed relatively worse on the NE metric. This suggests that different training setups were not effective in addressing the model's difficulty in determining whether the target has been successfully reached.

For the issue where tasks with fewer subtasks yield worse results, we consider the following reasons:

The first reason that the model performs worse in shorter tasks is that the model needs sufficient steps to warm up, i.e., setting up correct initial direction and locate targets. 
Though shorter tasks may seem easier, they leave fewer steps after initialization, making it more challenging for the agent to complete the task, thus leading to worse performance.

In our dataset analysis, as the number of subtasks in complex tasks increases, the average ground truth steps per subtask decrease (68.39, 53.30, 51.23), while the average maximum number of identical regions in the task increases (1.02, 1.89, 2.12), the probability of the agent observing critical information about subsequent subtasks during current subtasks increases(27.44\%, 42.11\%, 48.84\%). This suggests that with more subtasks, the probability of different subtask targets appearing in the same region increases, the average number of steps required for each subtask decreases, and the average difficulty of each subtask decreases. 

This is also reflected in the distances between subtask goals. In Table. \ref{tab:benchmark}, the navigation distance for the third subtask (i.e., the distance between the second and third goals) is always significantly smaller than the navigation distances at other stages. (4.74, 4.14 to 10.32, 11.47 and 9.74, 10.45) This indicates that in complex tasks with three or more subtasks, once the second goal is found, it becomes much easier to locate the third goal.

\begin{figure*}[!t] % 使用 figure* 环境占据双栏宽度
\centering
\begin{tcolorbox}[
    colback=gray!10,      % 设置背景色为10%的灰度
    colframe=black,       % 边框颜色为黑色
    width=\textwidth,     % 宽度为整栏
    boxrule=0.5pt,        % 边框线宽
    arc=4mm,              % 边框圆角半径
    left=2mm, right=2mm,  % 左右内边距
    top=2mm, bottom=2mm   % 上下内边距
]
\input{sec/prompts/prompt1}
\end{tcolorbox}
\vspace{-8pt}
\caption{$prompt_1$ for forward task generation}
\label{fig:prompt1}
\end{figure*}

\begin{figure*}[!t] % 使用 figure* 环境占据双栏宽度
\centering
\begin{tcolorbox}[
    colback=gray!10,      % 设置背景色为10%的灰度
    colframe=black,       % 边框颜色为黑色
    width=\textwidth,     % 宽度为整栏
    boxrule=0.5pt,        % 边框线宽
    arc=4mm,              % 边框圆角半径
    left=2mm, right=2mm,  % 左右内边距
    top=2mm, bottom=2mm   % 上下内边距
]
\input{sec/prompts/prompt2}
\end{tcolorbox}
\vspace{-8pt}
\caption{$prompt_2$ for backrward task generation}
\label{fig:prompt2}
\end{figure*}

\begin{figure*}[!t] % 使用 figure* 环境占据双栏宽度
\centering
\begin{tcolorbox}[
    colback=gray!10,      % 设置背景色为10%的灰度
    colframe=black,       % 边框颜色为黑色
    width=\textwidth,     % 宽度为整栏
    boxrule=0.5pt,        % 边框线宽
    arc=4mm,              % 边框圆角半径
    left=2mm, right=2mm,  % 左右内边距
    top=2mm, bottom=2mm   % 上下内边距
]
\input{sec/prompts/prompt3}
\end{tcolorbox}
\vspace{-8pt}
\caption{$prompt_3$ for MGDM.}
\label{fig:prompt3}
\end{figure*}

%% file: sec/prompts/prompt1.tex
\textbf{System}: \\
You are proficient in planning and design. You need to design a practical task consisting of multiple navigation-interaction subtasks based on the scene(which contain a large number of objects) and robot characteristics.
\\
\textbf{Rules}:\\
There are two part of the input: scene and robot. The scene includes objects in different regions of the scene. These objects serve as the foundation for generating your task, your task must be based on the scene. And the input robot includes the charactor of the robot, which is what you need to consider when you generate task.
\\
The task you generate should consist of the following three subtasks:
\\``````\\
- Move\_to("object\_region id"): Walk to an object in a region. PAY ATTENTION that "object\_region id" SHOULD be composed of the object and the ID of the corresponding region of the object in the input scene.\\
- Grab("object"): When the robot move to an object, it may need to grab it. To perform it, the robot need to move to the object first and make sure the robotic arm is empty. "object" should correspond to the object to be grab. PAY ATTENTION that "object" SHOULD be in the input scene.\\
- Release("object"): When the robot move to a place and had an object in hand, it can release the object in hands to the place. To perform it, the robot need to move to the place first and make sure the robotic arm is not empty. "object" should correspond to the object to be released. PAY ATTENTION that "object" SHOULD be in the input scene.\\
"""
\\
There are something you need to pay attention to:\\
``````\\
- Avoid words like "grab" and "release" in instructions.\\
- Pay attention that the objects in your task should be limited in 1 to 2 regions. And you should mention these regions in instruction. All of these regions SHOULD be in input scene.\\
- The task you generate should be similar to instructions like "Take an object in one region to a certain place in one region, then retrieve an object from that place."\\
- The object you take/grab should be portable.\\
- You need to generate a task consist of 4 to 6 subtasks.\\
- The region id in subtask "Move\_to" SHOULD be corresponded to the region in instruction.\\
"""\\
Your output should be a python dictionary which has two keys:\\
- dictionary["Task instruction"]: A conversational task instruction to describe the task.\\
- dictionary["Subtask list"]: A list of subtasks that make up the task.\\
Make sure the task instruction conversational enough, and the task should reasonable.
\\
\\
\textbf{Example}:\\
Here is an example of the INPUT and OUTPUT:\\
INPUT:\\
```\\
Scene: {"Region 0: Bedroom": ["bookshelf", "toy", "lamp", "whiteboard", "bed", "nightstand", "hanging clothes", "picture", "rug", "bag", "chest of drawers", "basket of something", "box", "clothes rack", "shoe"], "Region 3: Bathroom": ["door", "toilet", "bin", "toilet brush", "picture", "soap dispenser", "toilet paper", "sink", "sink cabinet"], "Region 4: Office": ["picture", "stationery", "statue", "ornament", "box", "folder", "printer", "stool", "bin", "plant", "computer", "mouse", "keyboard", "remote control", "chair", "water dispenser", "cushion", "desk", "light fixture"]}\\
Robot: "Spot in simulator is an agile, quadrupedal robot. 
Action: Spot support three kinds of action: move\_forward, turn\_left and turn\_right.
Sensors: Spot is equipped with three RGB cameras at a height of 0.5 meters in front, left and right to obtain embodied images in these three directions.
Mobility: Spot’s four-legged design allows it to navigate challenging terrains, including stairs, rocky surfaces, and cluttered environments.
Use: As spot is shaped like a dog, it can do some work for human in domestic scenes. It has a simple robotic arm, but is positioned lower (0.5 meters) and can perform some simple grabbing."\\
```\\
OUTPUT:
```
{"Task instruction": "take the bag in bedroom to the desk in office",
 "Subtask list": ["Move\_to('bag\_0')", "Grab('bag')", "Move\_to('desk\_4')", "Release('bag')"]}

%% file: sec/prompts/prompt2.tex
\textbf{System}:\\
You are good at guiding the way. Now you need to generate step-by-step navigation instructions based on some information.\\
\\
\textbf{Rules}:\\
The INPUT is a dictionary, the format is as follows:\\
```\\
\{"target": The target of navigation,\\
"step\_1": A dictionary containing the action of this step and the scene tags related to the environment of this step,\\
...\\
"step\_n": A dictionary containing the action of this step and the scene tags related to the environment of this step.\}\\
```\\
You need to combine the action of each step and the corresponding environment description scene tags based on the INPUT, and finally get a complete, easy-to-understand, and accurate navigation instruction. Please note the following requirements:

- You only need to select one most appropriate scene tag for each step to form a smooth and logical navigation instruction.\\
- The step corresponding to the “move\_forward” action should preferably select a specific scene from the scene tags.\\
- The step corresponding to the “turn\_left” and “turn\_right” actions should preferably select a specific object from the scene tags.\\
- The order of each action in the final instruction should be the same as the order of steps in the input, the total number of steps in the instruction should be the same as the number of steps in the INPUT, and the scene-related information involved should be in the scene tags of the corresponding action.\\
- The last step of the instruction should contain the navigation target, that is, the target in the INPUT.\\
- Your output should only be the instruction. The instruction should not contain words such as “step”.\\
\\
\textbf{Example}:\\
Here is an example:\\
```\\
INPUT:
\{'target': 'guitar case', 'step\_0': {'action': 'move\_forward', 'tags': ['store', 'retail', 'shopper', 'mall', 'fill']}, 'step\_1': {'action': 'make a left turn', 'tags': ['equipment', 'room', 'store', 'fill', 'retail']}, 'step\_2': {'action': 'move\_forward', 'tags': ['beam', 'retail', 'store', 'warehouse', 'room']}, 'step\_3': {'action': 'turn\_right', 'tags': ['retail', 'room', 'store', 'warehouse', 'fill']}, 'step\_4': {'action': 'move\_forward', 'tags': ['room', 'store', 'fill', 'shelf', 'retail']}, 'step\_5': {'action': 'turn\_left', 'tags': ['electronic', 'equipment', 'fill', 'room', 'store']}, 'step\_6': {'action': 'move\_forward', 'tags': ['retail', 'shelf', 'store', 'fill', 'room']}\}

OUTPUT:\\
Move forward through the store and make a left turn at the equipment, go ahead through the store, turn right at the retail shelf, move forward and turn left by the electronic equipment, finally go straight to the guitar case.\\
```\\
Please get the output according to the above prompts and the following INPUT. Think step by step.

%% file: sec/prompts/prompt3.tex
\textbf{"Task"}: "Instruction: Go to the locations to complete the given task. Task: ",\\
\textbf{"History"}: "Following is the History, which contains the visual information of your previous actions.",\\
\textbf{"Observation"}: "Following is the Candidate, which contains several directions you can go to at the current position, candidate (0) is stop.",\\
\textbf{"Output Hint"}: "Compare the History and Instruction to infer your current progress, and then select the correct direction from the candidates to go to the target location and finish the task."